\documentclass[10pt,journal,compsoc]{IEEEtran}
\ifCLASSOPTIONcompsoc
  \usepackage[nocompress]{cite}
\else
  \usepackage{cite}
\fi
\ifCLASSINFOpdf
  \usepackage[pdftex]{graphicx}
\else
\fi
\usepackage{amsmath}
\usepackage{bbm}
\usepackage{algorithmic}
\usepackage{array}
\ifCLASSOPTIONcompsoc
  \usepackage[caption=false,font=footnotesize,labelfont=sf,textfont=sf]{subfig}
\else
  \usepackage[caption=false,font=footnotesize]{subfig}
\fi
\usepackage{fixltx2e}
\usepackage{stfloats}
\usepackage{url}
\usepackage{amsmath,amsfonts}
\usepackage{array}
\usepackage{booktabs}
\usepackage{multirow}
\usepackage{hyperref}
\hypersetup{colorlinks,linkcolor={red},citecolor={cyan},urlcolor={magenta}} 
\usepackage{breakcites}
\usepackage[flushleft]{threeparttable}
\usepackage[misc,geometry]{ifsym}
\usepackage[table]{xcolor}

\usepackage{ragged2e}
\usepackage{xspace}
\usepackage{makecell} 

\definecolor{LightCyan}{rgb}{0.88,1,1}
\definecolor{Gray}{gray}{0.90}

\makeatletter
\DeclareRobustCommand\onedot{\futurelet\@let@token\@onedot}
\def\@onedot{\ifx\@let@token.\else.\null\fi\xspace}

\def\eg{\emph{e.g}\onedot} 
\def\ie{\emph{i.e}\onedot}

\makeatother

\begin{document}
% \title{PTTR++: Relational 3D Point Cloud Tracking with Transformer and Point-BEV Fusion}
% \title{PTTR++: Exploring Point-BEV Fusion for 3D Point Cloud Tracking with Transformer}
\title{Exploring Point-BEV Fusion for 3D Point Cloud Object Tracking with Transformer}

\author{Zhipeng Luo$\dagger$, Changqing Zhou$\dagger$, Liang Pan, Gongjie Zhang, Tianrui Liu,\\ 
Yueru Luo, Haiyu Zhao, Ziwei Liu, and Shijian Lu\ % <-this % stops a space
\IEEEcompsocitemizethanks{\IEEEcompsocthanksitem Zhipeng Luo, Changqing Zhou, Liang Pan, Gongjie Zhang, Yueru Luo, Ziwei Liu, and Shijian Lu are with the School of Computer Science and Engineering, Nanyang Technological University, Singapore. Tianrui Liu and Haiyu Zhao are with Sensetime Research.
\IEEEcompsocthanksitem E-mail: pan.liang@u.nus.edu{\tiny\,}, shijian.lu@ntu.edu.sg{\tiny\,}.
\IEEEcompsocthanksitem $\dagger$ denotes equal contribution}% <-this % stops an unwanted space
\thanks{Pre-print version. All rights reserved by the authors.}
}

\IEEEtitleabstractindextext{%
\begin{abstract}
\RaggedRight 
\justifying
With the prevalence of LiDAR sensors in autonomous driving, 3D object tracking has received increasing attention.
In a point cloud sequence, 3D object tracking aims to predict the location and orientation of an object in consecutive frames given an object template.
Motivated by the success of transformers, we propose \textbf{P}oint \textbf{T}racking \textbf{TR}ansformer (PTTR), which efficiently predicts high-quality 3D tracking results in a coarse-to-fine manner with the help of transformer operations.
PTTR consists of three novel designs. 
1) Instead of random sampling, we design \textit{Relation-Aware Sampling} to preserve relevant points to the given template during subsampling. 
2) We propose a \textit{Point Relation Transformer} for effective feature aggregation and feature matching between the template and search region.
3) Based on the coarse tracking results, we employ a novel \textit{Prediction Refinement Module} to obtain the final refined prediction through local feature pooling.
In addition, motivated by the favorable properties of the Bird's-Eye View (BEV) of point clouds in capturing object motion, we further design a more advanced framework named PTTR++, which incorporates both the point-wise view and BEV representation to exploit their complementary effect in generating high-quality tracking results. PTTR++ substantially boosts the tracking performance on top of PTTR with low computational overhead.
Extensive experiments over multiple datasets show that our proposed approaches achieve superior 3D tracking accuracy and efficiency. 
Code will be available at \url{https://github.com/Jasonkks/PTTR}.
\end{abstract}

}

\maketitle

\IEEEdisplaynontitleabstractindextext

% \fi
\IEEEpeerreviewmaketitle

\IEEEraisesectionheading{\section{Introduction}\label{sec:introduction}}


\begin{figure*}[t]
    \centering
    \includegraphics[width=1.0\linewidth]{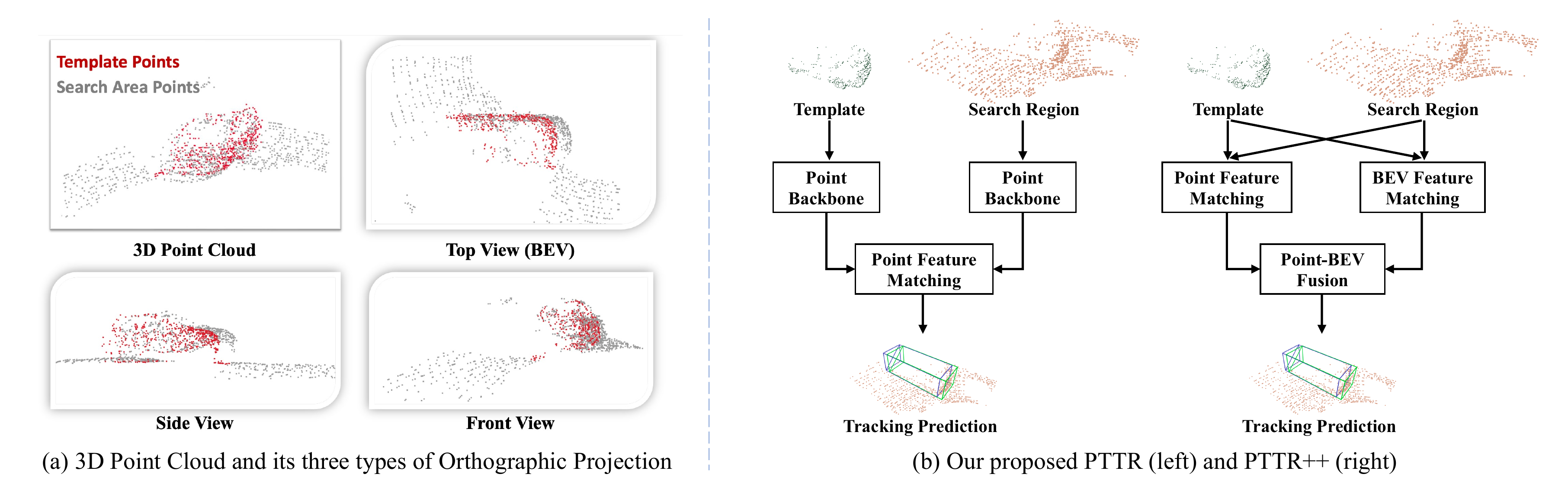}
    \vspace{-8mm}
    \caption{(a) The Bird's-Eye View (BEV) densifies the point cloud and filers out the noises in the height dimension, while most object movements occur in the horizontal plane. Template points from the previous frame are overlaid with search area points from the current frame for illustration. (b) PTTR++ builds on top of PTTR by performing BEV feature matching in parallel with the existing point feature matching to exploit the complementary information of the two representations.}
    \label{fig:view_comparison}
    % \vspace{-3mm}
\end{figure*}

\IEEEPARstart{W}{ith} the rapid development of 3D sensors in the past decade, solving various vision problems \cite{qi2017pointnet,qi2017pointnet++,20203dssd,shi2019pointrcnn,luo2021unsupervised,xiao2021synlidar,xiao2022unsupervised,landrieu2018large,ren2022benchmarking,pan2021variational} with point clouds has attracted increasing attention due to the huge potential in applications such as autonomous driving, motion planning, and robotics. As a long-standing research problem in computer vision, object tracking with point clouds has also drawn wide research interests. 
3D object tracking aims to determine the object pose and position of the tracked object in consecutive frames in a point cloud sequence given the object template in the first frame.
% 3D object tracking aims to detect not only object poses and positions in each frame but also object motion trajectories across consecutive frames. 
However, 3D tracking still faces a number of open and challenging problems such as LiDAR point cloud sparsity, random shape incompleteness, texture feature absence, etc.

% Object tracking is a long-standing research problem in computer vision.
% With the rapid development of 3D sensors in the past decade, 3D tracking with scanned point clouds becomes increasingly important in various tasks such as 3D environment understanding, motion prediction, and path planning. Taking autonomous driving as an example, 

% \begin{figure}[t]
%     \centering
%     \includegraphics[width=1.0\linewidth]{figures/teaser_5.pdf}
%     % \vspace{-5mm}
%     \caption{
%     \textbf{(a)} 3D point cloud object tracking aims to track the target object based on a given template point cloud.
%     \textbf{(b)} PTTR outperforms existing approaches by large margins on KITTI tracking dataset \cite{kitti}.
%     \textbf{(c)} We visualize our tracking results over consecutive frames, including ``100 Frames'', ``200 Frames'' and ``300 Frames'', which
%     demonstrate the robustness of PTTR for long-term tracking.}
%     \label{fig:teaser_success}
%     % \vspace{-3mm}
% \end{figure}

Existing 3D object tracking approaches can be largely categorized into two groups: multi-object tracking (MOT) and single-object tracking (SOT). 
MOT methods~\cite{weng20203d,wang2021joint,weng2020graph,yin2021center} generally adopt a detect-to-track strategy by first detecting objects in each frame and then matching the detections across consecutive frames based on the estimated location or speed.
In contrast, SOT methods only process a subset of the point cloud scene, which usually come with much lower computational costs and higher throughput.
We study SOT in this work, and our objective is to estimate the location and orientation of a single object in the search frame given an object template.

The pioneer 3D SOT method SC3D~\cite{2019sc3d} first generates a series of candidates given the last location of a specific object, and then makes predictions by selecting the best-matched candidate in the latent space.
However, it is not end-to-end trainable and suffers from low inference speed as SC3D relies on a large number of candidates.
Differently, P2B~\cite{2020p2b} explores not relying on many candidates by first using cosine similarity to fuse features of the search region with the template, and then adopting the prediction head of VoteNet~\cite{qi2019deep} to generate the final prediction.
Following P2B, SA-P2B~\cite{zhou2021structure} adds an extra auxiliary network to predict the object structure. 
In a similar framework, 3D-SiamRPN~\cite{fang20203d} uses a cross-correlation module for feature matching and an RPN head for final prediction.
These methods~\cite{2019sc3d, 2020p2b, zhou2021structure} essentially perform a linear matching process between features in the search domain and the template, which is typically not robust against various challenges in 3D observation, such as random noise, sparsity, and occlusions.
Moreover, the inclusion of complex prediction heads as in detection models \cite{qi2019deep} severly limits their tracking speed, which is a crucial factor for real-time applications.

% In this work, we design Point Tracking TRansformer (PTTR), a novel tracking paradigm that exploits transformer operations to achieve high-quality 3D object tracking.
% % in a coarse-to-fine fashion.
% Specifically, we first propose \textbf{PTTR}, a transformer-based model that generates tracking predictions in a coarse-to-fine manner. PTTR first extracts point features from the template and search area individually using the PointNet++ \cite{qi2017pointnet++} backbone.
In this work, we design \textbf{P}oint \textbf{T}racking \textbf{TR}ansformer (PTTR), a novel tracking paradigm that achieves high-quality 3D object tracking in a coarse-to-fine fashion.
Specifically, PTTR first extracts point features from the template and search area individually using the PointNet++ \cite{qi2017pointnet++} backbone. 
To alleviate the point sparsity issue, which is common for objects at distance, we propose a sampling strategy termed \textit{Relation-Aware Sampling}, which can preserve more points that are relevant to the given template by leveraging the relation-aware feature similarities between the search and the template. 
We then propose a novel \textit{Point Relation Transformer} equipped with \textit{Relation Attention Module} to match search and template features and generate a coarse prediction based on the matched feature. PRT first utilizes a self-attention operation to adaptively aggregate point features for the template and the search area individually, and then performs feature matching with a cross-attention operation.
Moreover, we propose a lightweight \textit{Prediction Refinement Module} to refine the coarse prediction with local feature pooling. We highlight that PTTR achieves good efficiency despite the prediction refinement process.

% KITTI tracking dataset \cite{kitti} has been widely adopted in 3D tracking evaluations. However, it has clear constraints including limited sample size and highly imbalanced class distributions. We create a new point cloud tracking benchmark named Waymo SOT Dataset based on Waymo Open Dataset \cite{waymo}, which has a large sample size as well as balanced class distributions. The new benchmark is thus complementary to the KITTI tracking dataset by offering more holistic and comprehensive evaluations to the 3D tracking research community. 
% We evaluate PTTR on both datasets, and 

Other than the commonly used point-wise representation in 3D SOT, another type of point cloud representation that has been widely adopted in various 3D perception tasks \cite{qi2016volumetric,zhou2018voxelnet,yan2018second} and achieves great success is the voxel-based view. 
Specifically, the raw point clouds are rasterized into voxel cells, hence leading to a structured 3D format.
% , which is favorable to efficient local geometrical feature extraction and relationship modeling.
One special form of voxel representation is the Bird's-Eye View (BEV), where the height dimension is suppressed and the point cloud is converted to 2D feature maps.
The BEV representation was first introduced in 3D object detection \cite{lang2019pointpillars} and is known for its computation efficiency.
From the inspection of point cloud tracklets, we find that BEV has a significant potential to benefit 3D tracking.
As shown in Fig.~\ref{fig:view_comparison}\textcolor{red}{(a)}, BEV could better capture motion features compared to the other two views (\ie, Side View and Front View) since object movements largely occur in the horizontal plane in scenarios such as autonomous driving.
By compressing the 3D point clouds, BEV naturally filters out the noises in the height dimension, which makes it promising for identifying object motions in a tracking sequence.
Nonetheless, voxelization and BEV representations inevitably cause information loss, which could lead to inaccurate localization, especially for smaller objects (\eg, pedestrians). Therefore, it is intuitive to explore the complementary effect of point-wise and BEV representations for the tracking task to exploit their merits.
% Therefore, one promising direction would be exploring the complementary effect of both point-based and BEV representations for the tracking task.

In light of the above motivations, we further propose a 3D tracking framework named PTTR++ on top of PTTR to exploit the complementary effect between point features and BEV features. As shown in Fig.~\ref{fig:view_comparison}\textcolor{red}{(b)}, PTTR++ consists of two parallel feature matching branches for the point-wise view and BEV, respectively, where the point branch is directly inherited from PTTR and each branch matches template and search features with PRT independently. Subsequently, \textit{Cross-view Feature Mapping} is performed to establish the mapping between point and BEV features. Finally, we propose \textit{Selective Feature Fusion}  to adaptively fuse the mapped point and BEV features for generating the tracking prediction. With the help of the proposed Point-BEV fusion strategy, PTTR++ substantially outperforms PTTR in terms of tracking accuracy with low computational overhead. 
While we mainly develop our method based on PTTR, we highlight that the proposed Point-BEV fusion is a generic approach which can be easily integrated with other 3D trackers. We experiment on another recently published method M\textsuperscript{2}-Track \cite{zheng2022beyond}, which is a contemporary work to our conference paper \cite{zhou2022pttr}, to validate the wide applicability of our proposed fusion approach.
% We also show that the proposed fusion is a generic approach, which can be integrated with other point cloud tracking methods to boost the performance.

% \begin{figure}[t]
%     \centering
%     \includegraphics[width=1.0\linewidth]{figures/point cloud views.png}
%     % \vspace{-3mm}
%     \caption{3D point cloud and its three types of orthographic projection. The Bird's-Eye View (BEV) densifies the point cloud and filers out the noises in the vertical dimension while most of the object movements occur in the horizontal plane. Template points from the previous frame are overlaid with search area points from the current frame for illustration.}
%     \label{fig:view_comparison}
%     % \vspace{-3mm}
% \end{figure}

% The contributions of this work are threefold. 
In summary, the contributions of this work are:
1) We design PTTR, a transformer-based 3D point cloud object tracking method, which performs tracking in a coarse-to-fine manner. PTTR consists of a few novel designs, including Relation-Aware Sampling for preserving template-relevant points, Point Relation Transformer for effective feature aggregation and matching, and a lightweight pooling-based refinement module.
2) We further propose PTTR++ on top of PTTR to explore the complementary effect of point-wise and BEV representations for improved tracking performance. PTTR++ employs Cross-view Feature Mapping and Selective Feature Fusion for effective Point-BEV fusion.
3) Comprehensive evaluations over multiple benchmarks show that our proposed methods achieve state-of-the-art performance with competitive inference speed.

This work is an extension of our conference paper \cite{zhou2022pttr} published at \textit{CVPR'\,2022}. Compared with the conference version, we incorporate the following new contents. 1) We propose a more advanced tracking framework PTTR++ by incorporating Point-BEV fusion with PTTR to achieve state-of-the-art performance. 2) We show that the proposed fusion strategy is a generic approach that can be applied to boost the performance of other tracking methods. 3) We provide a more comprehensive performance evaluation by reporting experimental results on multiple benchmarks and conducting extensive ablation studies.

\section{Related Works}

\noindent\textbf{2D Object Tracking.} Majority of the recent 2D object tracking methods follow the Siamese network paradigm, composed of two CNN branches with shared parameters that help project inputs into the same feature space. \cite{tao2016siamese} employs Siamese network to learn a generic matching function for different objects. At inference, a bunch of candidates are used to match the original target and the one that matches best is chosen as the prediction. \cite{bertinetto2016fully} proposes a fully-convolutional Siamese architecture to locate the target object in a larger search area. 
% The output is actually a heatmap with its maximum response representing the result. 
\cite{guo2017learning} proposes a dynamic Siamese network that learns to transform the target appearance and suppress the background. \cite{li2018high,li2019siamrpn++} apply Siamese network to extract features and use pair-wise correlation separately for classification branch and regression branch of the region proposal network (RPN). 2D tracking approaches are not directly applicable to point clouds as they are driven by 2D CNN architectures and they are not designed to address the unique challenges of 3D tracking. 

\smallskip
\noindent\textbf{3D Object Tracking.} 3D object tracking can be roughly divided into two categories: multi-object tracking (MOT) and single-object tracking (SOT). Most MOT approaches adopt a detect-to-track strategy and mainly focus on data association \cite{xiang2015learning,karunasekera2019multiple}. \cite{weng2019baseline} first proposes a 3D detection module to provide the 3D bounding boxes, then use 3D Kalman Filter to predict current estimation, and match them using Hungarian algorithm. \cite{wang2021joint} proposes to use GNN to model relationships among different objects both spatially and temporally, while \cite{yin2021center} uses a closest distance matching after speed compensation.
SOT methods focus on tracking a single object given a template. Most of the existing 3D SOT approaches utilize the Siamese network architecture to match the template and search area. As the pioneer work, SC3D \cite{2019sc3d} proposes to match feature distance between candidates and target and regularize the training using shape completion. P2B \cite{2020p2b} matches search and template features with cosine similarity and employs Hough Voting \cite{qi2019deep} to predict the current location. SA-P2B \cite{zhou2021structure} proposes to learn the object structure as an auxiliary task. 3D-SiamRPN \cite{fang20203d} uses a RPN \cite{ren2015faster} head to predict the final results. BAT \cite{zheng2021box} encodes box information in Box Cloud to incorporate structural information. MLVSNet \cite{wang2021mlvsnet} proposes to perform multi-level Hough voting for aggregating information from different levels. PTT \cite{shan2021ptt} proposes a Point-Track-Transformer module to weight features' importance.  Most existing SOT methods either use cosine similarity or cross-correlation to match the search and template features, which are essentially linear matching processes and cannot adapt to complex situations where random noise and occlusions are involved. Moreover, the use of detection model prediction heads leads to high computation overheads. Our proposed PTTR addresses the above limitations. In particular, one recent work \cite{zheng2022beyond} proposes a new motion-centric tracking paradigm, which directly predicts the motion state based on overlapped point cloud frames. We empirically show that our proposed Point-BEV fusion method in PTTR++ is also applicable to this paradigm.

\smallskip
\noindent\textbf{Vision Transformers.} Transformer \cite{vaswani2017attention} was first proposed as an attention-based building block in machine translation to replace the RNN architecture. Recently, a number of works \cite{liu2021swin,chu2021twins,carion2020end,zhu2020deformable,xie2021segformer,zhang2022meta,zhang2022accelerating} apply transformer on 2D vision tasks and achieve great success. Most of these attempts divide the images into overlapping patches and then regard each patch as a token to further apply the transformer architecture.
In the 3D domain, PCT \cite{guo2021pct} generates positional embedding using 3D coordinates of points and adopts transformer with an offset attention module to enrich features of points from its local neighborhood. Point Transformer \cite{zhao2021point} adopted vectorized self-attention network \cite{zhao2020exploring} for local neighbours and designed a Point Transformer layer that is order-invariant to suit point cloud processing. \cite{engel2021point} proposes SortNet to gather spatial information from point clouds, which sorts the points by learned scores to achieve order invariance. All of these works focus on shape classification or part segmentation tasks.
The attention mechanism in transformer offers correlation modeling with global receptive field, which makes it a good candidate for the 3D tracking problem where feature matching is required. We propose a novel transformer-based module to utilize the attention mechanism for feature aggregation and matching.

\smallskip
\noindent\textbf{Multi-view Feature Fusion.} Point cloud can be represented in different views and some recent methods \cite{liu2019point,shi2020pv,xu2021rpvnet,noh2021hvpr,wei2021pv,zhang2020deep} attempt to fuse two or more different views for better performance for tasks such as classification and detection.
PVCNN \cite{liu2019point} proposes to use convolution on voxelized features to replace the expensive ball-query operations in PointNet++ \cite{qi2017pointnet++} and improve the efficiency.
PV-RCNN \cite{shi2020pv} proposes to integrate voxel-based and point-based networks with two aggregation modules: voxel-to-keypoint encoding for fusing multi-scale semantic features into points and keypoint-to-grid RoI pooling for aggregating local context information.
RPVNet \cite{xu2021rpvnet} designs a network with three branches for voxel, point, and range image, respectively. The voxel and range image views are converted to the point view for multi-view feature propagation at different stages. The fused point features are then projected back to other views. A gating mechanism is used to measure the importance of different views and adaptively aggregate features.
FusionNet \cite{zhang2020deep} proposes a deep fusion architecture where neighborhood aggregation is applied to process point and voxel features, and inner-voxel aggregation is used for point-level and voxel-level feature interaction.
To our best knowledge, it is the first attempt to adapt multi-view fusion to the 3D tracking task where feature matching is involved. 
Our proposed PTTR++ differs from these methods by using the BEV representation and an attention-based selective fusion method based on cross-view mapping.

\begin{figure*}[t]
    \centering
    \includegraphics[width=1.0\linewidth]{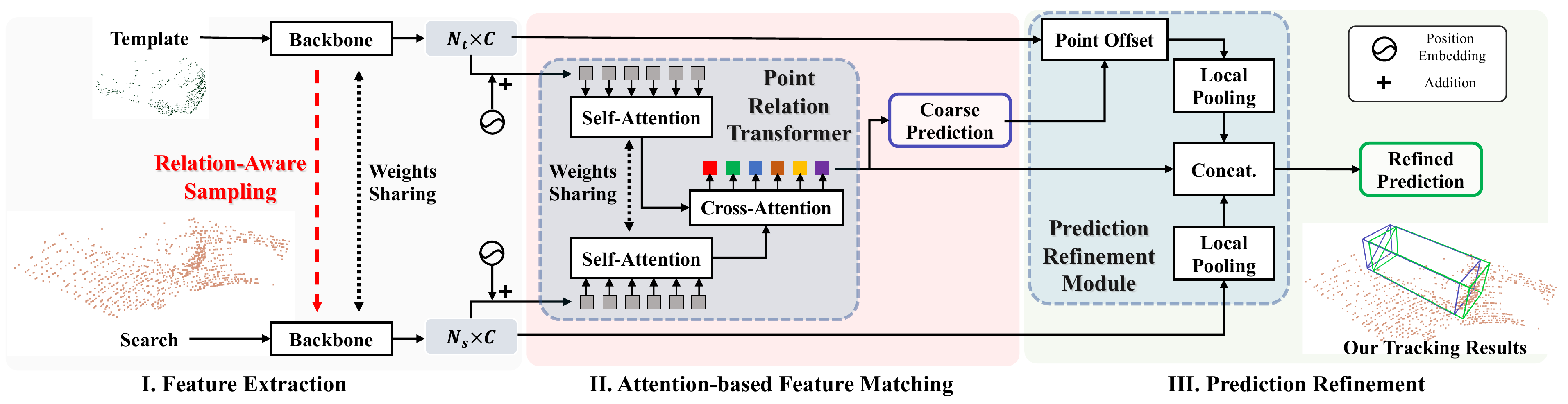}
    \vspace{-2mm}
    \caption{\textbf{Overview of our proposed PTTR.} The network mainly consists of three parts: feature extraction, attention-based feature matching and prediction refinement. The backbone is used to extract features from input point clouds. We modify PointNet++ \cite{qi2017pointnet++} with our proposed \textit{Relation-Aware Sampling} to help select more positive points from the search area. For feature matching, we propose \textit{Point Relation Transformer} equipped with \textit{Relation Attention Module} to match search and template features. In the prediction stage, we propose a \textit{Prediction Refinement Module} to generate predictions in a coarse-to-fine manner.}
    \label{fig:overview}
    % \vspace{-3mm}
\end{figure*}

% \input{latex/sections/3_framework}
% \section{Method} \label{method}
\section{PTTR: 3D Point Cloud Tracking with Transformer} \label{sec:pttrv1}
\subsection{Overview} \label{sec:overview}
Given a 3D point cloud sequence, 3D object tracking aims to estimate the object location and orientation in each point cloud observation, \ie, the search point cloud $P^s \in \mathbb{R}^{N_s \times 3}$, by predicting a bounding box conditioned on a template point cloud $P^t \in \mathbb{R}^{N_t \times 3}$. 
To this end, we propose PTTR, a novel coarse-to-fine framework for 3D object tracking.
As shown in Fig.~\ref{fig:overview}, PTTR performs 3D point cloud tracking with three main stages: 1) Feature Extraction (Sec.~\ref{sec:sample}), 2) Attention-based Feature Matching (Sec.~\ref{sec:transformer}), and 3) Prediction Refinement (Sec.~\ref{sec:refinement}).

\smallskip
\noindent\textbf{Feature Extraction.}
Following previous methods \cite{2020p2b, 2019sc3d, zhou2021structure, fang20203d}, we employ PointNet++~\cite{qi2017pointnet++} as the backbone to extract multi-scale point features from the template and the search.
% To alleviate this point sparsity issue, 
However, important information loss may occur during random subsampling in the original PointNet++.
We therefore propose a novel 
\textit{Relation-Aware Sampling} method
to preserve more points relevant to the given template by leveraging relation-aware feature similarities.

\smallskip
\noindent\textbf{Attention-based Feature Matching.}
Different from previous methods that often use cosine similarity \cite{2019sc3d,2020p2b,zhou2021structure} or linear correlation \cite{fang20203d} for matching the template and the search, 
we utilize novel attention operations and propose \textit{Point Relation Transformer} (PRT).
% in Sec.~\ref{sec:transformer}.
PRT first utilizes a self-attention operation to adaptively aggregate point features for the template and the search area individually, and then performs feature matching with cross-attention.
The coarse prediction is generated based on the output of PRT.
% exploits cross-contextual information with a cross-attention operation.
% By matching features

% \noindent\textbf{Coarse-to-Fine Tracking Prediction.}
\smallskip
\noindent\textbf{Prediction Refinement.}
% With the help of PRT, we generate the coarse predictions that 
The coarse prediction is further refined with a lightweight \textit{Prediction Refinement Module}, which results in a coarse-to-fine tracking framework.
% introduced in Sec.~\ref{sec:refinement}.
Based on the coarse predictions, we first conduct a Point Offset operation for seed points from the search to estimate their corresponding seed points in the template.
Afterwards, we employ a Local Pooling operation for the seed points from both point clouds respectively, and then concatenate the pooled features with the matched features from PRT for estimating our final prediction.

% }

\subsection{Relation-Aware Feature Extraction} \label{sec:sample}

As one of the most successful backbones, PointNet++~\cite{qi2017pointnet++} introduces a hierarchical architecture with multiple distance-farthest point sampling (D-FPS) and ball query operations, which effectively exploits multi-scale point features.
Most existing 3D object tracking methods~\cite{2020p2b, zhou2021structure, fang20203d} use PointNet++ for feature extraction.
However, it has a non-negligible disadvantage for object tracking: the D-FPS sampling strategy used in PointNet++ tends to generate random samples that are uniformly-distributed in the euclidean space, which often leads to important information loss during the sampling process.
In particular, the search point cloud often has a much larger size than the template, and therefore D-FPS sampling inevitably keeps a substantial portion of background points and leads to sparse point distribution for the object of interest, which further challenges the subsequent template searching using feature matching.
To alleviate this problem, previous methods use either random point sampling~\cite{2020p2b, zhou2021structure} or feature-farthest point sampling (F-FPS)~\cite{20203dssd}.
However, the problem of substantial foreground information loss during sampling is not fully resolved.

\smallskip
\noindent\textbf{Relation-Aware Sampling.}
In contrast, we propose to use a novel sampling method dubbed \textit{Relation-Aware Sampling} (RAS) to preserve more points relevant to the given template by considering relational semantics.
Our key insight is that the region of interest in the search point cloud should have similar semantics with the template.
% Therefore, those points with higher semantic feature similarities between the template point clouds and the search are more probably to be foreground points.
Therefore, points in search area with higher semantic feature similarities to the template points are more likely to be foreground points.
Specifically, given the template point features $\mathbf{X}^t \in \mathbb{R}^{N_t \times C}$ and search area point features $\mathbf{X}^s \in \mathbb{R}^{N_s \times C}$, we first calculate the pairwise point feature distance matrix $\mathbf{D} \in \mathbb{R}^{N_s \times N_t}$:
\begin{equation}
    \mathbf{D}_{ij} = ||\mathbf{x}^s_i - \mathbf{x}^t_j||_2,\;\;  \forall\, \mathbf{x}^s_i \in \mathbf{X}^s,\;\; \forall\, \mathbf{x}^t_j \in \mathbf{X}^t,
\end{equation}
where $||\cdot||_2$ denotes L2-norm, and $N_s$ and $N_t$ are the current number of points from the search area and the template, respectively. 
Afterwards, we compute the minimum distance $\mathbf{V} \in \mathbb{R}^{N_s}$ by considering the distance between each point from search and its nearest point from template in the feature space:
\begin{equation}
    \mathbf{V}_i = \min_{j=1}^{N_t} (\mathbf{D}_{ij}),\;\;  \forall\, i \in \{1,2,...,N_s\}.
\end{equation}

\begin{figure}[t]
    \centering
    \includegraphics[width=1.0\linewidth]{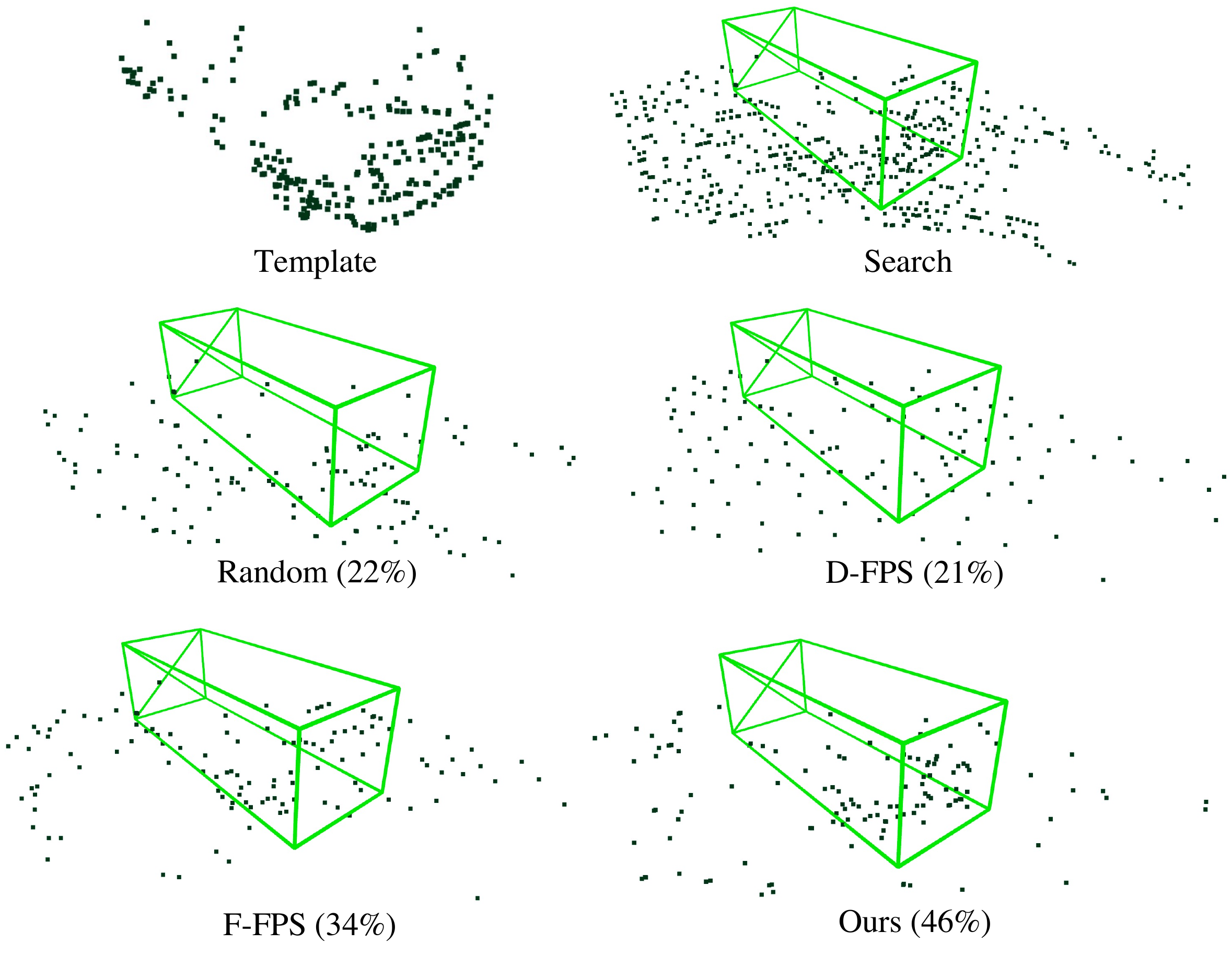}
    % \vspace{-6mm}
    \caption{\textbf{Comparison of sampling methods}. We show the sampled points from the search area using different sampling methods. Our proposed sampling method preserves the most number of points belonging to the object. The percentage in the figure represents the ratio of positive points to all sampled points.}
    \label{fig:sample_method}
% \vspace{-3mm}
\end{figure}

Following previous methods~\cite{2020p2b, 2019sc3d, zhou2021structure, fang20203d},
% During tracking inference, 
we update the template point cloud for each frame by using the tracking result from the previous search point cloud.
% For a few challenging cases, 
% However,
% it is possible to obtain low-quality template point clouds for challenging cases, which may mislead the RAS sampling. 
In the case that low-quality tracking predictions are encountered in difficult situations, the newly formed templates might mislead the RAS and lead to unfavorable sampling results. Moreover, the inclusion of background information offers useful contextual information for the localization of the tracked object. To improve the robustness of the sampling process, we adopt a similar strategy as in \cite{20203dssd} to combine our proposed RAS with randomly sampling. In practice, we sample half of the points with RAS, while the rest of the points are obtained via random sampling. We show the effects of different sampling approaches in Fig.~\ref{fig:sample_method}. It can be observed that the proposed sampling method can preserve the most object points.

\subsection{Relation-Enhanced Feature Matching} \label{sec:transformer}
% \pl{
% Under the Tracking-by-Detection paradigm, 
Existing 3D object tracking methods perform feature matching between the search point cloud and template by using cosine similarity \cite{2019sc3d,2020p2b,zhou2021structure} or linear correlation \cite{fang20203d}.
On the other hand, motivated by the success of various attention-based operations for computer vision applications~\cite{pan2021variational,zhao2021point,vaswani2017attention,guo2021pct}, we strive to explore attention-based mechanism for 3D tracking, which can adapt to different noisy point cloud observations. Although PTT \cite{shan2021ptt} utilizes transformer in their model, they still match template and search point cloud by cosine similarity and the transformer module is only used for feature enhancement.
% }

\begin{figure}[t]
    \centering
    \includegraphics[width=0.6\linewidth]{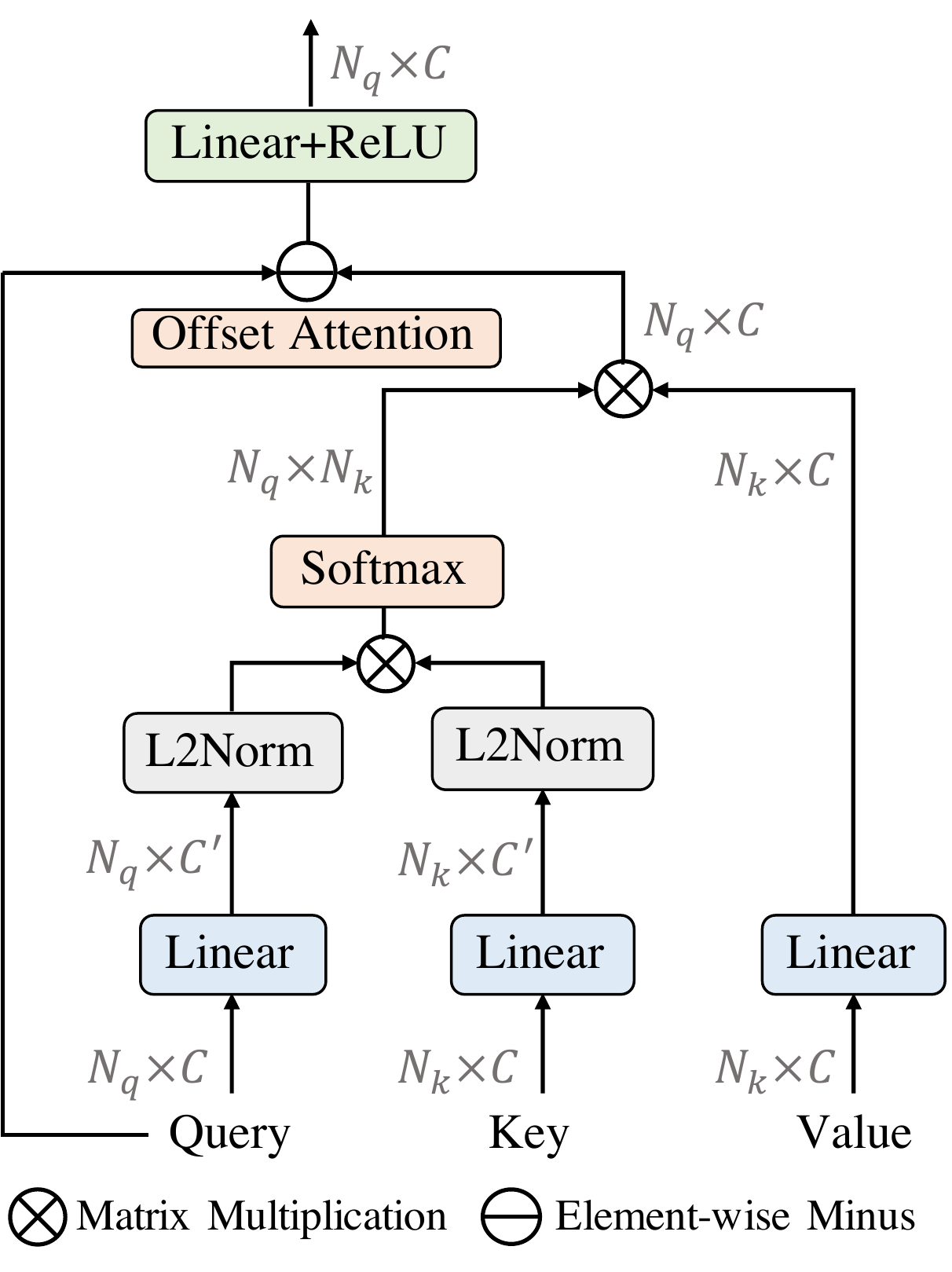}
    % \vspace{-3mm}
    \caption{\textbf{Architecture of the Relation Attention Module (RAM).} 
    RAM first projects query, key and value into a latent feature space and then estimates the attention matrix by multiplying the L2-normalized query and key features. The attention matrix is then applied on the value feature to obtain the attention product before the offset attention~\cite{guo2021pct} operation and Linear and ReLU layers for injection of non-linearity.
    % We further input the matrix-multiplication results of attention score and value into offset attention~\cite{guo2021pct} and finally a linear+ReLU layer.
    }
    \label{fig:relation attention}
% \vspace{-3mm}
\end{figure}

% \pl{
\smallskip
\noindent\textbf{Relation Attention Module.}
Inspired by recent works studying feature matching~\cite{wang2021transformer, wang2018cosface,wang2017normface,chen2020simple}, we propose the \textit{Relation Attention Module} (RAM) (shown in Fig.~\ref{fig:relation attention}) to adaptively aggregate features by predicted attention weights.
Firstly, RAM employs linear projection layers to transform the input feature vectors ``Query'', ``Key'' and ``Value''.
% into latent space.
Instead of naively calculating the dot products between ``Query'' and ``Key'', RAM predicts the attention map by calculating the cosine distances between the two sets of L2-normalized feature vectors.
With the help of L2-normalization, RAM can prevent the dominance of a few feature channels with extremely large magnitudes. 
Subsequently, the attention map is normalized with a Softmax operation.
In order to sharpen the attention weights and meanwhile reduce the influence of noise~\cite{guo2021pct}, we employ the offset attention to predict the final attention map by subtracting the query features with the previously normalized attention map.
Consequently, the proposed RAM can be formulated as:
\begin{align}
    \text{Attn}(\mathbf{Q}, \mathbf{K}, \mathbf{V}) = \phi\big(\mathbf{Q}& - \text{softmax}(\mathbf{A})\cdot(W_v \mathbf{V})\big),
\end{align}
where $\phi$ represents the linear layer and ReLU operation applied to the output features, the attention matrix $\mathbf{A} \in \mathbb{R}^{N_q \times N_{k}}$ is obtained by:
\begin{align}
    \mathbf{A} = \overline{\mathbf{Q}} \cdot \overline{\mathbf{K}}^{^\top},\;
    \overline{\mathbf{Q}} = \frac{W_q \mathbf{Q}}{||W_q \mathbf{Q}||_2}, \; 
    \overline{\mathbf{K}} = \frac{W_k \mathbf{K}}{||W_k \mathbf{K}||_2},
\end{align}
where $\|\cdot\|_2$ is the L2-norm,
$\mathbf{Q}, \mathbf{K}, \mathbf{V}$ represent the input ``Query'', ``Key'' and ``Value'' respectively, and $W_q$, $W_k$ and $W_v$ denote the corresponding linear projections.
% }

\smallskip
\noindent\textbf{Point Relation Transformer.}
By incorporating RAM,
% Specifically, 
we propose the \textit{Point Relation Transformer} (PRT) module
% that incorporates a novel attention mechanism to 
to adaptively exploit the correlations between point features for context enhancement.
The PRT module firstly performs a self-attention operation for the search and the template features, respectively. 
Subsequently, PRT employs a cross-attention operation for gathering cross-contextual information between the two point clouds.
Both operations use global attentions, where all input point feature vectors are considered as tokens.
Formally, PRT is formulated as:
\begin{align}
    \bar{\mathbf{X}}^s & = \text{Attn}(\mathbf{X}^s), \;\;\text{and}\;\;
    \bar{\mathbf{X}}^t  = \text{Attn}(\mathbf{X}^t), \\
    \hat{\mathbf{X}}^s & = \text{Attn}(\bar{\mathbf{X}}^s, \bar{\mathbf{X}}^t, \bar{\mathbf{X}}^t), 
\end{align}
where $\text{Attn}(\mathbf{Q}, \mathbf{K}, \mathbf{V})$ denotes our proposed \textit{Relation Attention Module}, $\hat{\mathbf{X}}^s$ denotes the matched features, and $\bar{\mathbf{X}}^s$ and $\bar{\mathbf{X}}^t$ denote the enhanced search and template features respectively.
By using global self-attention, the exploited features can obtain a global understanding for the current observation.
Note that the self-attention use the same point features as $\mathbf{Q}, \mathbf{K}, \mathbf{V}$, and both self-attention operations share weights so as to project the search and template features into the same latent space.
Thereafter, the cross-attention performs pairwise matching between query tokens $\bar{\mathbf{X}}^s$ and key tokens $\bar{\mathbf{X}}^t$, which exploits cross-contextual information for $\hat{\mathbf{X}}^s$ by capturing correlations between the two sets of point features.
% to capture 
% the correlation between the two sets of points.
Based on the relation enhanced point features $\hat{\mathbf{X}}^s$, we can generate our coarse prediction results for 3D object tracking.

\subsection{Coarse-to-Fine Tracking Prediction} \label{sec:refinement}
Majority of the existing point tracking approaches adopt prediction heads of detection models to generate the predictions, e.g. P2B \cite{2020p2b} adopts the clustering and voting operations of VoteNet \cite{qi2019deep} and 3D-SiamRPN \cite{fang20203d} uses a RPN \cite{ren2015faster,li2018high} head. 
% Despite their effectiveness, these prediction heads introduce extra computation overheads and limit the inference speed. 
However, these prediction heads introduce extra computation overheads, which largely limits their efficiency.
To circumvent this issue, 
% we propose a novel prediction framework 
% that efficiently performs high-quality 3D tracking in a coarse-to-fine manner.
we propose a novel coarse-to-fine tracking framework.
The coarse prediction $\mathbf{Y}^c$ is predicted by directly regressing the relation enhanced features $\hat{\mathbf{X}}^s$ from the proposed PRT module with Multi-Layer-Perceptron (MLP).
Remarkably, $\mathbf{Y}^c$ provides faithful tracking predictions for most cases, and also surpasses the tracking performance of SoTA methods.

\smallskip
\noindent\textbf{Prediction Refinement Module.}
To further refine the tracking predictions, we propose a lightweight \textit{Prediction Refinement Module} (PRM) to predict our final predictions $\mathbf{Y}^f$ based on $\mathbf{Y}^c$.
Specifically, we use the sampled points from the search point cloud as seed points, and then we estimate their correspondences in the template by using an offset operation for $\mathbf{Y}^c$. 
Afterwards, we encode local discriminative feature descriptors for the seed points from both sources, which is achieved by using Local Pooling operations for grouped neighboring point features.
The neighboring features are grouped by using ball-query operations with a fixed radius $r$. 
In the end, we concatenate $\hat{\mathbf{X}}^s$ with the pooled features from the source and the target, based on which we generate the final prediction $\mathbf{Y}^f$:  
\begin{align}
    \mathbf{Y}^f = \gamma([\mathbf{F}^s, \mathbf{F}^t, \hat{\mathbf{X}}^s]),
\end{align}
where $\mathbf{F}^s$ and $\mathbf{F}^t$ are the pooled features from search and template respectively, $[\cdot]$ denotes concatenation operation, and $\gamma$ represents the MLP networks. 
We highlight that even with the refinement stage, our proposed method still achieve competitive inference speed thanks to the lightweight design. 

% \subsection{Training Loss}
\smallskip
\noindent\textbf{Training Loss.}
Our PTTR is trained in an end-to-end manner.
The coarse prediction $\mathbf{Y}^c$ and the final prediction $\mathbf{Y}^f$ are in the same form that each contains a classification component $\mathbf{Y}_{cls} \in \mathbb{R}^{N_s\times1}$ and a regression component $\mathbf{Y}_{reg} \in \mathbb{R}^{N_s\times4}$, where $N_s$ denotes the number of sampled points from the search area. $\mathbf{Y}_{cls}$ predicts the objectiveness of each point, and $\mathbf{Y}_{reg}$ consists of the predicted offsets along each axis $\{\Delta{x}, \Delta{y}, \Delta{z}\}$ with an additional rotation angle offset $\Delta{\theta}$. 
% The predictions of both stages consist of classification as well as regression and we use binary cross entropy (BCE) and mean square error (MSE) for their loss computations respectively:
For each prediction, we use a classification loss $\mathcal{L}_{cls}$ defined by binary cross entropy, and a regression loss $\mathcal{L}_{reg}$ computed by mean square error.
Consequently, our overall loss function is formulated as:
\begin{align}
    \mathcal{L}_{total} =& \mathcal{L}_{cls}(\mathbf{Y}^c_{cls}, \mathbf{Y}^{gt}_{cls}) + \mathcal{L}_{reg}(\mathbf{Y}^c_{reg}, \mathbf{Y}^{gt}_{reg}) +\\ &\lambda\big(\mathcal{L}_{cls}(\mathbf{Y}^f_{cls}, \mathbf{Y}^{gt}_{cls}) + \mathcal{L}_{reg}(\mathbf{Y}^f_{reg}, \mathbf{Y}^{gt}_{reg})\big),
\end{align}
where $\mathbf{Y}^{gt}_{(\cdot)}$ denotes the corresponding ground truth, and $\lambda$ is a weighting parameter.

\begin{figure*}[t]
    \centering
    \includegraphics[width=1.0\linewidth]{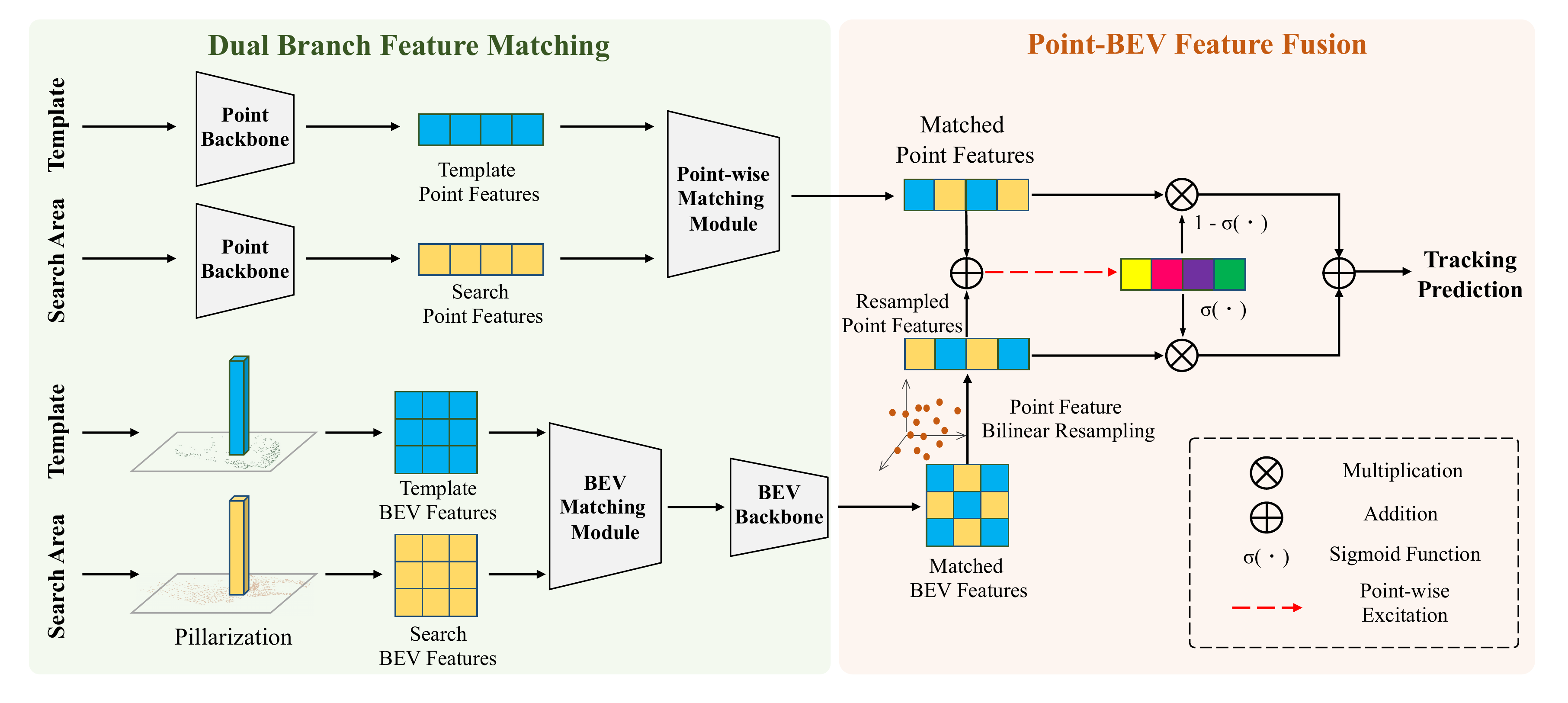}
    \vspace{-6mm}
    \caption{\textbf{Overview of our proposed PTTR++}. 
    % The network consists of two parallel branches to exploit the complementary information in the point-wise and BEV representations. 
    PTTR++ extends from PTTR with an additional BEV branch in parallel with the existing point branch to exploit the complementary information in the point-wise and BEV representations. In Dual Branch Feature Matching, PTTR++ performs template-search feature matching based on the two views, respectively. In Point-BEV Feature Fusion, the matched BEV features are first projected to the point domain to fuse with the point branch counterpart for generating the final prediction.}
    \label{fig:pttrv2}
    % \vspace{-3mm}
\end{figure*}

\subsection{Implementation Details} \label{sec:implementation_details_v1}
\noindent\textbf{Template and Search Region.} 
During training, we use the ground truth bounding box to crop the point cloud to form the template. 
In order to simulate the disturbances the model might encounter, we add random distortions to augment the bounding boxes with a range of [-0.3 to 0.3] along x, y, z axes. 
For both training and testing, we extend the box with a ratio of 0.1 to include some background points. 
We enlarge the template bounding box by 2 meters in all directions to form the search region.

\smallskip
\noindent\textbf{Model Details.} 
We use PointNet++\cite{qi2017pointnet++} with 3 set-abstraction layers as the backbone. 
The radius of these SA layers is set to 0.3, 0.5, 0.7 meters, respectively.
% We use one transformer layer for the PRT module.
% as we have experimented on stacking more transformer layers and no notable performance gain is observed. 
In the first stage, we use a 3-layer MLP for classification and regression, respectively. 
Each layer is followed by a BN \cite{ioffe2015batch} layer and an ReLU \cite{agarap2018deep} activation layer. 
In PRM, the Local Pooling is conducted with a ball-query operation and a grouping operation \cite{qi2017pointnet++} with a radius of 1.0 meter.
After pooling, we obtain the concatenated features that are fed into a 5-layer MLP for generating the final predictions.

\section{PTTR++: Exploring Point-BEV Fusion for 3D Point Cloud Tracking} \label{sec:pttrv2}

\subsection{Overview}
As introduced in Sec. \ref{sec:introduction}, motivated by the favorable properties of the Bird's-Eye View in capturing object motion and its potential in complementing the point-wise representation, we further propose a new 3D tracking framework on top of PTTR and name it as PTTR++, which explores Point-BEV fusion for point cloud tracking. As shown in Fig.~\ref{fig:pttrv2}, PTTR++ consists of two stages, namely Dual Branch Feature Matching (in Sec.~\ref{method_matching}) and Point-BEV Feature Fusion (in Sec.~\ref{method_fusion}). 
In the first stage, PTTR++ performs template-search feature matching for the point and BEV features independently with transformer operations.
In the second stage, it first maps the features from one branch to another, and then adaptively fuse the mapped futures to generate the tracking prediction. Note that we do not include the Prediction Refinement Module in PTTR++ as the focus is the exploration of the synergy of the two point cloud representations. Our empirical results demonstrate that PTTR++ achieves significantly improved performance and competitive efficiency without incorporating PRM thanks to the strong complementary effect of the two representations.

\subsection{Dual Branch Feature Matching} \label{method_matching}
To exploit the complementary information in the point-wise view and its corresponding BEV representation, we introduce two parallel feature matching branches, namely the \textit{point branch} and the \textit{BEV branch}. 
Each branch consists of a backbone network that extracts features for the template and search area respectively, as well as a transformer-based matching module that performs template and search feature matching.
% In addition, we compare and discuss different matching strategies (\ie, early matching vs late matching) for each branch.
In this stage, the two branches encode features individually.

The point branch is inherited from PTTR that PRT is used to perform feature matching between template and search point features as introduced in Sec. \ref{sec:pttrv1}. We denoted the matched point features as $F^P$. 
For the BEV branch, we first perform the pillarization process to convert the raw point cloud into BEV features. 
As Fig.~\ref{fig:pttrv2} illustrates, we first divide the x-y plane into evenly spaced gird to form the pillars.
For each pillar, the points that fall inside the pillar are first augmented with the arithmetic mean of the point coordinates $(x_c, y_c, z_c)$ and the distance from the pillar center $(x_p, y_p)$. 
Following BAT \cite{zheng2021box}, we additionally add object size information to the points by attaching the ground truth bounding box size $(w, h, l)$ to each point. 
As a result, together with the original point coordinates $(x, y, z)$, the point features have 11 dimensions in total. 
We then use a linear layer to encode the pillars followed by a max-pooling operation to obtain the pillar features of shape $(P, C)$, where $P$ is the number of pillars and $C$ is the number of feature channels. 
Note that the pillarization process is efficient for sparse point clouds since only non-empty pillars need to be processed. 
Subsequently, the pillars are scattered back to their original location inside the feature maps to obtain template and search BEV features of shape $(H, W, C)$, where $H$ and $W$ indicate the 2D grid size.
By considering each element inside the feature maps as a token, similar to the point branch, we employ the proposed PRT to match BEV features.
Differently, we use the sinusoidal function~\cite{vaswani2017attention} to generate positional embeddings for each BEV feature token.
The output of the matching module is then processed by a BEV backbone consisting of convolution layers to obtain the final matched BEV features $F^B$.
As shown in Fig. \ref{fig:pttrv2}, the BEV branch differs from the point branch that the BEV matching module is positioned before the BEV backbone which consists of convolution layers that downsample the feature maps. It allows the feature matching to be performed at higher resolution, which potentially leads to more accurate localization. We empirically verify that the proposed setting brings improved tracking performance. We refer the readers to our ablation studies in Sec. \ref{sec:ablation_pttr++} for details.

\subsection{Point-BEV Feature Fusion} \label{method_fusion}

To fuse the matched features from the point and BEV branches, we first leverage \textit{Cross-view Feature Mapping} (CFM) to transform features from one branch to the other (\eg, from the point branch to the BEV branch).
% operations, which estimates the correspondences between the two sets of features.
% Both two directions, point-to-BEV and BEV-to-point mapping operations, are introduced.
Afterwards, we propose a \textit{Selective Feature Fusion} (SFF) operation to adaptively fuse the mapped cross-view features.
In this stage, we focus on combining the advantages of the point-based view and the BEV representation for tracking prediction.

\noindent\textbf{Cross-view Feature Mapping.} 
% To fuse the matched features from both branches, we first perform a feature mapping process to project the features from one view to the other. Here we introduce both point-to-BEV and BEV-to-point mapping operations. 
CFM can be performed in both directions, namely Point-to-BEV mapping and BEV-to-Point mapping, and we introduce both of them here.
The Point-to-BEV mapping is achieved via grid-based average pooling, which is similar to the pillarization process introduced in Sec.~\ref{method_matching}. 
Based on the corresponding point coordinates $\{s_k\}$ of the point features $\{F_k^P\}$ as well as the range and grid size of the target BEV representation, the Point-to-BEV mapping can be described as:
\begin{align}
    F_{h, w}^{P \to B} = \frac{1}{N_{h,w}}\sum_{k=1}^K \mathbbm{1}[s_k \in \mathcal{P}_{h,w}]F_k^P
\end{align}
where $h$ and $w$ index the target BEV feature maps $F^{P \to B}$, $\mathcal{P}$ is the set of pillars corresponding to the grid, $K$ is the number of points, $N_{h,w}$ is the number of points within $\mathcal{P}_{h,w}$, and $\mathbbm{1}$ is the indicator function. 
% On the other hand, 

The BEV-to-Point mapping can be regarded as the reserve process of Point-to-BEV mapping. 
Instead of assigning the same features to all the points that fall inside the same pillar, we use bilinear interpolation to resample point features from the BEV features based on their corresponding x-y coordinates. 
% We name this mapping process point feature bilinear resampling and 
The resampled point features are denoted as $F^{B\to P}$.

\begin{figure}[t]
    \centering
    \includegraphics[width=1.0\linewidth]{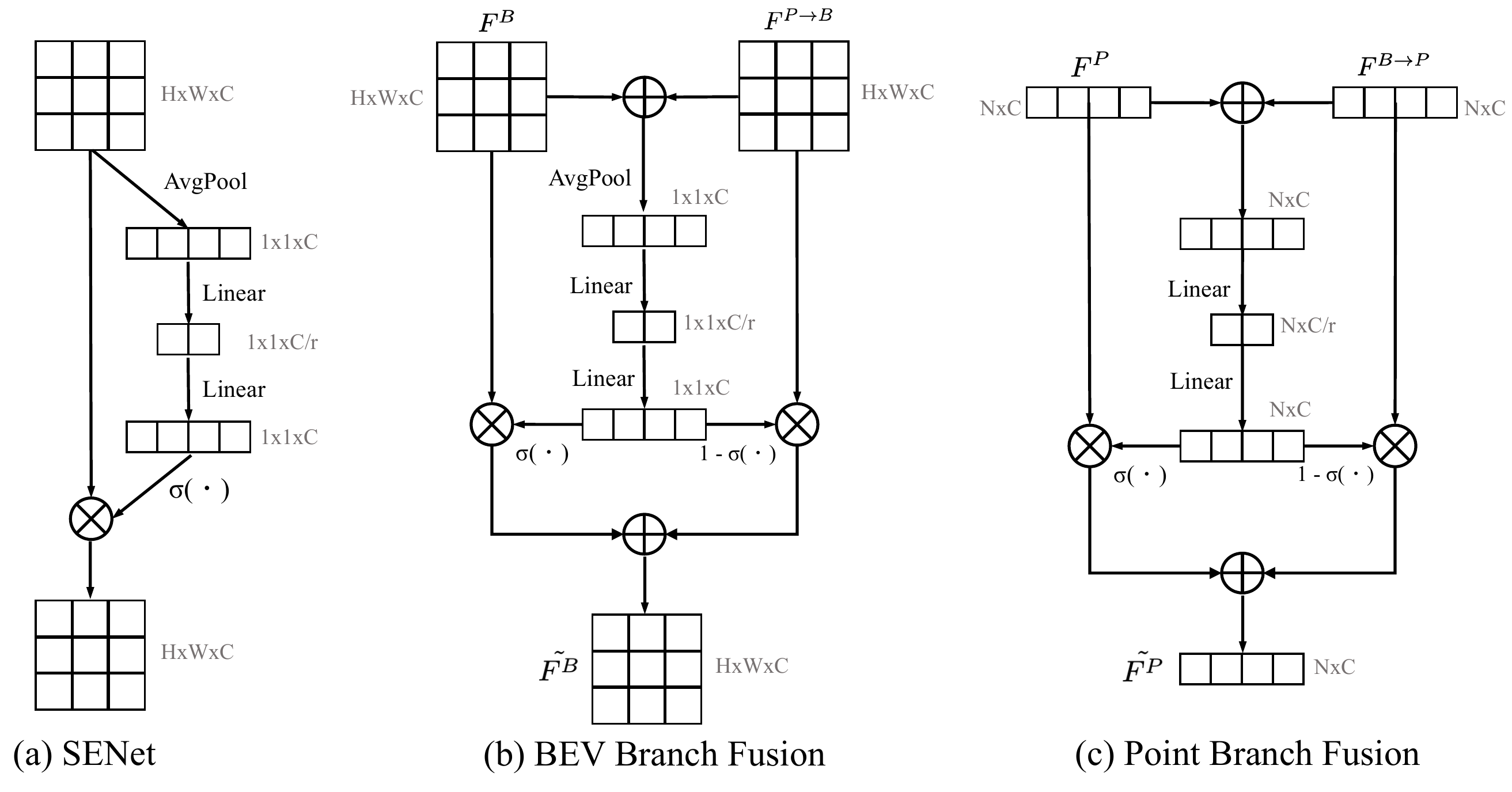}
    % \vspace{-5mm}
    \caption{
    \textbf{Selective feature fusion}. The BEV branch adopts global channel-wise re-weighting to fuse two sets of features, while the point branch performs re-weighting for each individual point. r denotes the reduction ratio as defined in \cite{hu2018squeezesenet}.
    }
    % \vspace{-3mm}
    \label{fig:attention}
\end{figure}

\smallskip
\noindent\textbf{Selective Feature Fusion.}
% After cross-view feature mapping, one intuitive and straightforward approach to fusing both features is through simple addition. However, since different views contain complementary information, it is sub-optimal to treat both representations equally in the fusion process. Therefore, 
After obtaining the mapped features from the 3D point cloud and BEV representations, 
% we propose to fuse the two views through an attention mechanism to assign different weights to each view. 
we propose to adaptively fuse the two sets of features following the similar spirit of SENet~\cite{hu2018squeezesenet}.
% Our method is inspired by SENet \cite{hu2018squeezesenet} which recalibrates channel-wise features to emphasize more informative features and suppress the others. 
As Fig.~\ref{fig:attention}\textcolor{red}{(a)} illustrates, SENet works on 2D image features and it first performs the ``squeeze" operation by downsampling the feature maps via global average pooling, followed by the ``excitation" operation which generates channel-wise attention weights with linear layers and the sigmoid function. 
Finally, the attention weights are broadcast and multiplied with the input features as a gating mechanism. 
Our method differs from SENet that two sets of features from the point and BEV branches are involved. 
In order to generate meaningful attention weights that can effectively select informative features from both inputs, we first sum the features from both branches before the squeeze and excitation operations that generate the attention weights. 
By taking the BEV branch as an example (Fig.~\ref{fig:attention}\textcolor{red}{(b)}), we can formulate this process as:
\begin{equation}
    w = \sigma(\gamma(\text{AvgPool}(F^B + F^{P \to B})))
\end{equation}
where $\gamma$ and $\sigma$ represent linear layers and the sigmoid function, and $w$ denotes the attention weights. We then multiply $w$ and $1-w$ to both input features before the summation so that the fusion process essentially acts as a self-gating mechanism to adaptively select useful information from both branches: 

\begin{equation}
    \Tilde{F^B} = w F^B + (1-w) F^{P \to B}
\end{equation}
where $\Tilde{F^B}$ denotes the fused features on the BEV branch. On the other hand, for the point branch, as points are unstructured and different points might have distinct feature responses, making it unsuitable to apply a global channel-wise attention weight. We instead remove the global average pooling step to perform channel-wise re-weighting for each individual point as shown in Fig.~\ref{fig:attention}\textcolor{red}{(c)}. The rest of the fusion process is similar to the BEV branch and the fused point features are denoted by $\Tilde{F^P}$. For simplicity, we name the attention mechanism for the BEV branch as global attention and the point-wise version (for the point branch) as point-wise attention.

In our empirical analysis, we conduct experiments to evaluate feature fusion on both branches and select BEV-to-Point mapping and fusion on the point branch as our proposed method (as illustrated in Fig.~\ref{fig:pttrv2}) as it achieves more competitive and balanced performance. Please refer to the experimental results in Table \ref{tab:ablation_fusion} for details.

\smallskip
\noindent\textbf{Training Loss.}
We apply the same prediction head and training loss as PTTR on the fused point features $\Tilde{F^P}$. Note that the Prediction Refinement Module proposed in PTTR is omitted to reduce the computational cost. In addition, we also apply a prediction head on the matched BEV features $F^B$ to obtain the BEV prediction $Y^B$ to allow for additional supervision on the BEV branch. The overall loss function is computed as:
\begin{equation}
    \begin{split}
        \mathcal{L} = \mathcal{L}_{BCE}(Y_{cls}^P,\, &\Bar{Y}_{cls}^P) + \alpha \mathcal{L}_{MSE}(Y_{reg}^P,\, \Bar{Y}_{reg}^P) \\ 
        &+ \mathcal{L}_{BCE}(Y_{cls}^B,\, \Bar{Y}_{cls}^B) + \beta \mathcal{L}_{MSE}(Y_{reg}^B,\, \Bar{Y}_{reg}^B)\;,
    \end{split}
\end{equation}
where $Y^P$ and $Y^B$ denote the predictions of both branches, $\Bar{Y}$ indicates the ground truth, and $\alpha$ and $\beta$ are the loss coefficients. During inference, prediction is only generated on the point branch.

\subsection{Implementation Details}
We mostly follow the settings of PTTR as introduced in Sec. \ref{sec:implementation_details_v1}. For the additional BEV branch, we use a point cloud range of [(-4.8, 4.8), (-4.8, 4.8), (-1.5, 1.5)] meters and a pillar size of 0.3 meter for smaller objects including car, van, pedestrian, and cyclist. For large objects such as truck, trailer, and bus, we adopt a range of [(-12.0, 12.0), (-12.0, 12.0), (-4.0, 4.0)] meters and a pillar size of 0.5 meter. The point features are projected to 64 dimensions using one-layer MLP after pillarization. In the BEV backbone network, we use 3 convolution blocks with [128, 128, 256] channels respectively and each block has a down-sample factor of 2. In attention-based feature fusion, we follow SENet \cite{hu2018squeezesenet} and use a reduction ratio $r$ of 16. For loss computation, we set the coefficients $\alpha$ and $\beta$ as 100 and 2, respectively.

\section{Experiments}

\subsection{Experimental Settings}

\noindent\textbf{Datasets.}
We evaluate our proposed method on the most commonly used KITTI \cite{kitti} and nuScenes \cite{caesar2020nuscenes} SOT datasets. For a fair comparison on the KITTI dataset, we follow the data split specified in \cite{2019sc3d,2020p2b} and use scenes 0-16 for training, 17-18 for validation, and 19-20 for testing. For the nuScenes dataset, we follow the implementation of \cite{zheng2021box} and report the performance of five categories including Car, Pedestrian, Truck, Trailer, and Bus. The nuScenes dataset only provides annotation for 1 in 10 consecutive frames, and the annotated frames are defined as keyframes. The evaluation is performed on keyframes only.

\smallskip
\noindent\textbf{Evaluation Metrics.}
We follow existing works \cite{2020p2b,hui20213dv2b} and use \textit{Success} and \textit{Precision} as defined in one pass evaluation (OPE) \cite{kristan2016novel} as the evaluation metrics. Specifically, \textit{Success} measures the IoU between the predicted box and the ground truth, while \textit{Precision} computes the AUC of the distance between prediction and ground truth box centers. We report the performance of each object category as well the average over all classes.

\smallskip
\noindent\textbf{Training and Testing.} 
For the KITTI tracking dataset, we train the model for 160 epochs with a batch size of 64. We use Adam optimizer \cite{kingma2014adam} with an initial learning rate of 0.001 and reduce it by 5 every 40 epochs. 
% For the Waymo SOT Dataset, we train the model for 80 epochs with the same initial learning rate and reduce the learning rate every 20 epochs. 
For nuScenes \cite{caesar2020nuscenes}, we train the model for 30 epochs with a batch size of 128. We set the initial learning rate to 0.001 and divide it by 5 every 6 epochs.
During testing, we use the previous prediction result as the next template. 
In line with \cite{2019sc3d,2020p2b}, we use the ground truth bounding box as the first template.

\begin{table*}[t]
\setlength{\tabcolsep}{20pt}
\small
\caption {\textbf{Performance comparison on the KITTI dataset}. Success / Precision are used for evaluation. * indicates results produced based on the official implementation. M\textsuperscript{2}-Track++ denotes M\textsuperscript{2}-Track with our proposed Point-BEV fusion. `Improvement' refers to performance gain introduced by our proposed Point-BEV fusion.} \label{tab:kitti_experiment} 
\vspace{-4mm}
\begin{center}\setlength{\tabcolsep}{4pt}{
\scalebox{1.0}{
\begin{tabular}{c|c|c|c|c|c|c}
\toprule[1.2pt]
% {\bf Method} & {\bf Car } & {\bf Pedestrian} & {\bf Van} & {\bf Cyclist} & \bf Average\\ 
\multirow{2}{*}{Method} & Car & Pedestrian &  Van & Cyclist &  Average & Average \\
 & 6424 & 6088 & 1248 & 308 & by Class & by Frame \\
% {Method} & {Car} & {Pedestrian} & {Van} & {Cyclist} &  Average by class & Average by instance \\
\midrule
\small SC3D \cite{2019sc3d} &  \;41.3 / 57.9\; &  \;18.2 / 37.8\; &  \;40.4 / 47.0\; &  \;41.5 / 70.4\; &  \;35.4 / 53.3\; &  \;31.2 / 48.5\; \\
\small P2B \cite{2020p2b} &  56.2 / 72.8 &  28.7 / 49.6 &  40.8 / 48.4 &  32.1 / 44.7 &  39.5 / 53.9 &  42.4 / 60.0 \\
\small 3D-SiamRPN \cite{fang20203d} &  58.2 / 76.2 &  35.2 / 56.2 &  45.7 / 52.9 &  36.2 / 49.0 &  43.8 / 58.6 &  46.6 / 64.9\\
\small SA-P2B \cite{zhou2021structure} &  58.0 / 75.1 &  34.6 / 63.3 &  51.2 / 63.1 &  32.0 / 43.6 &  44.0 / 61.3 &  46.7 / 68.2 \\
 MLVSNet \cite{wang2021mlvsnet} &  56.0 / 74.0 &  34.1 / 61.1 &  52.0 / 61.4 &  34.3 / 44.5 &  44.1 / 60.3 & 45.7 / 66.7 \\
\small BAT \cite{zheng2021box} &  60.5 / {77.7} &  42.1 / 70.1 &  52.4 / {67.0} &  33.7 / 45.4 &  47.2 / 65.1 &  51.2 / 72.8 \\
PTT \cite{shan2021ptt} &  {67.8} / {81.8} &  44.9 / 72.0 &  43.6 / 52.5 &  37.2 / 47.3 &  48.4 / 63.4 & 55.1 / 74.2\\
LTTR \cite{cui20213dlttr} &  65.0 / 77.1 &  33.2 / 56.8 &  35.8 / 45.6 &  66.2 / 89.9 &  50.0 / 67.4 &  48.7 / 65.8 \\
V2B \cite{hui20213dv2b} &  70.5 / 81.3 &  {48.3 / 73.5} &  50.1 / 58.0 &  40.8 /49.7 &  {52.4 / 65.6} &  58.4 / 75.2 \\
\midrule
M\textsuperscript{2}-Track \cite{zheng2022beyond} * & 64.8 / 79.2 & 57.6 / 83.6  & 50.4 / 64.8 & 74.9 / 93.3 & 61.9 / 80.2 & 60.6 / 80.1 \\
M\textsuperscript{2}-Track++ (Ours) * & 67.0 / 80.3 & {61.3} / {88.6} &  {58.3} / {75.4} &  {75.5} / {93.6} &  {65.5} / {84.5} &  {64.0} / {83.7} \\
Improvement & +2.2 / +1.1 & +3.7 / +5.0 & +7.9 / +10.6 & +0.6 / +0.3 & +3.6 / +4.3 & +3.4 / +3.6 \\
\midrule
PTTR (Ours) &  {65.2} / 77.4 &  {50.9 / 81.6} &  {52.5} / 61.8 &  {65.1 / 90.5} &  {58.4 / 77.8} &  57.9 / 78.1 \\
% PTTR-Lite & 69.8 / 80.7 & 50.2 / 79.4 &  51.5 / 58.5 &  {73.7} / {93.0 } &   61.3 / 77.9 &  59.8 / 78.4 \\
PTTR++ (Ours) &  {73.4} / {84.5} &  {55.2} / {84.7} &  {55.1} / 62.2 &  71.6 / 92.8 &  {63.8} / {81.0} &  {63.9} / {82.8} \\
Improvement & +8.2 / +7.1 & +4.3 / +3.1 & +2.6 / +0.4 & +6.5 / +2.3 & +5.4 / +3.2 & +6.0 / +4.7 \\
\bottomrule[1.2pt]
\end {tabular}}
}
\vspace{-2mm}
\end{center}
\end {table*}

\begin{table*}[t]
\setlength{\tabcolsep}{20pt}
\small
\caption {\textbf{Performance comparison on the nuScenes dataset}. The experimental results of comparing methods are reprinted from \cite{zheng2021box}. Success / Precision are used for evaluation. `Improvement' refers to performance gain introduced by our proposed Point-BEV fusion.} \label{tab:nuscenes_exp} 
\vspace{-6mm}
\begin{center}\setlength{\tabcolsep}{2pt}{
\scalebox{1.0}{
\begin{tabular}{c|c|c|c|c|c|c|c}
\toprule[1.2pt]
\multirow{2}{*}{Method} & Car & Pedestrian & Truck & Trailer &  Bus &  Average & Average\\ 
 & 64159 & 33227 & 13587 & 3352 & 2953 & by Class & by Frame \\
\midrule
\small SC3D \cite{2019sc3d} &\;  22.31 / 21.93\; & \; 11.29 / 12.65\; & \; 30.67 / 27.73\; & \; 35.28 / 28.12\;  & \;  29.35 / 24.08\;  &\;  25.78 / 22.90\; &\;  20.70 / 20.20 \; \\
\small P2B \cite{2020p2b} &  38.81 / 43.18 &  28.39 / 52.24 &  42.95 / 41.59 &  48.96 / 40.05 &  32.95 / 27.41 &  38.41 / 40.90 &  36.48 / 45.08 \\
\small BAT \cite{zheng2021box} &  40.73 / 43.29 &  28.83 / 53.32 &  45.34 / 42.58 &  {52.59}  / {44.89} &  35.44 / 28.01 &  40.59 / 42.42 &  38.10 / 45.71 \\
% V2B \cite{hui20213dv2b} &  - &  - &  - &  - &  - & \\
M\textsuperscript{2}-Track \cite{zheng2022beyond} &  55.85 / 65.09 & 32.10 / 60.92 &  57.36 / 59.54 &  57.61 / 58.26 &  51.39 / 51.44 & 50.86 / 59.05 & 49.23 / 62.73 \\
\midrule
% PTTR-Lite &  51.24 / 56.42 &  34.98 / 50.29 &  50.05 / 48.87 &  43.71 / 35.23 &  50.42 / 47.74 &  46.08 / 47.71 &  46.26 / 52.98 \\
PTTR (Ours) & 51.89 / 58.61 & 29.90 / 45.09 & 45.30 / 44.74 & 45.87 / 38.36 & 43.14 / 37.74 &  43.22 / 44.91 & 44.50 / 52.07   \\
PTTR++ (Ours) & 59.96 / 66.73 &	32.49 / 50.50 &	59.85 / 61.20 & 	54.51 / 50.28 & 53.98 / 51.22 & 52.16 / 55.97 & 51.86 / 60.63 \\
Improvement & + 8.07 / +8.12 & +2.59 / +5.41 & +14.55 / +16.46 & +8.64 / +11.94 & +10.84 / +13.48 & +8.94 / +11.06 & +7.36 / +8.56 \\
\bottomrule[1.2pt]
\end {tabular}}
}
% \vspace{-4mm}
\end{center}
\end {table*}

% \subsection{3D Tracking on KITTI Tracking Dataset}
\subsection{Benchmarking Results}

\noindent\textbf{Results on KITTI.}
% for fair comparison. 
%  and use scenes 0-16 for training, 17-18 for validation, and 19-20 for testing.
% The tracklets are generated by concatenating all frames where the instance appears.
We compare PTTR and PTTR++ with existing state-of-the-art 3D SOT methods. In particular, to validate that the proposed Point-BEV fusion in PTTR++ is a generic approach, we integrate it with a recently published method M\textsuperscript{2}-Track \cite{zheng2022beyond} and denote it as M\textsuperscript{2}-Track++. M\textsuperscript{2}-Track is a contemporary work to our conference version, which differs from existing matching-based methods that it predicts the motion state based on overlapped point cloud frames. We refer the readers to \cite{zheng2022beyond} for details. We incorporate the proposed Point-BEV fusion to the targetness prediction stage that voxelization is performed on the overlapped point cloud input to form BEV features, which is fused with point features to generate the segmentation scores. As reported in Table~\ref{tab:kitti_experiment}, PTTR outperforms all existing matching-based methods while PTTR++ further boosts the average performance by a large margin of +6.0/+4.7 in success and precision, respectively. On top of M\textsuperscript{2}-Track, our simple integration of the proposed Point-BEV fusion strategy also brings a notable improvement of +3.4/+3.6 on average, showing the wide applicability of the proposed approach.

% As reported in Table~\ref{tab:kitti_experiment}, PTTR surpasses the previous state-of-the-art method by a significant margin of 8.4 and 10.4 in terms of average Success and Precision. 
% % Notably, our method achieves larger performance gains for Pedestrian and Cyclist, which are known to be difficult categories due to their small object size and limited number of points.
% In particular, PTTR significantly outperforms SA-P2B on the challenging categories, Pedestrian and Cyclist.
% As shown in Fig.~\ref{fig:ours vs p2b}, we compare the proposed PTTR against P2B \cite{2020p2b} over two pedestrian point cloud sequences. 
% P2B often makes wrong predictions when multiple instances are close, while PTTR is able to generate stable and reliable predictions.
% The impressive performance gains are largely attributed to the proposed PRT and RAS modules, which will be further explained in the ablation studies in Sec.~\ref{sec:ablations}.
% Moreover, we visualize the coarse and refined predictions in Fig.~\ref{fig:second stage}.
% % Moreover, we visualize the effectiveness of the coarse-to-fine tracking framework,
% The coarse predictions are further corrected in the refinement stage, especially when point sparsity or large movements are present. 
% It demonstrates that our proposed prediction refinement via local feature pooling is able to adapt to challenging situations and generate robust predictions. 

\smallskip
\noindent\textbf{Results on nuScenes.} Table \ref{tab:nuscenes_exp} compares the tracking performance on the nuScenes dataset. Our proposed PTTR again demonstrates competitive performance and outperforms existing matching based methods. In addition, PTTR++ further improves the tracking accuracy as compared to PTTR by a significant margin of 8.94/11.06 in average success and precision and achieves state-of-the-art performance. We find that the proposed Point-BEV fusion is especially beneficial to larger objects such as truck and bus. 

\smallskip
\noindent\textbf{Inference Time.}
Efficiency has always been an important aspect of object tracking due to its time-sensitive applications. We compare the inference time of our method with existing methods with available open-source implementations on the KITTI dataset. We run all the models on the same machine and report the inference time in Table \ref{tab:speed_comparsion}. It can be observed that our methods have comparable inference time as existing methods. In particular, PTTR++ only adds a small proportion of inference time on top of PTTR despite having an extra BEV branch. This is because the BEV branch is lightweight thanks to the efficient pillarization process and convolution operations as opposed to the expensive point grouping operation.

% speed comparison
\begin{table*}[t]
\setlength{\tabcolsep}{10pt}
\small
\caption{\textbf{Inference time comparison on the KITTI dataset.}} \label{tab:speed_comparsion} 
\vspace{-4mm}
\begin{center}\setlength{\tabcolsep}{4pt}{
\scalebox{0.9}{
\begin{tabular}{c|c|c|c|c|c|c}
\toprule
Model & P2B & BAT & M\textsuperscript{2}-Track & M\textsuperscript{2}-Track++ (Ours) & PTTR (Ours) & PTTR++ (Ours) \\
\midrule
%  & Early & 69.8 / 80.7 & 50.2 / 79.4 &  51.5 / 58.5 & \textbf{73.7} / 93.0 &  61.3 / 77.9 & \textbf{16.4 ms} \\
% Mid & Early & 70.5 / 81.3 & 53.1 / 81.9 & 53.7 / 61.1 & 73.1 / \textbf{93.3}  & 62.6 / 79.4 & 20.3 ms\\
Average Performance & 42.4 / 60.0 & 51.2 / 72.8 & 60.6 / 80.1 & 64.0 / 83.7 & 57.9 / 78.1 & 63.9 / 82.8 \\
\midrule
Inference Time (ms) & 23.5 & 21.4 & 17.5 & 20.8 & 19.4 & 23.1 \\
\bottomrule[1.2pt]
\end {tabular}}
}
\vspace{-3mm}
\end{center}
\end {table*}

% \begin{table}[h]
% \caption{\textbf{Inference Time.}}
% \label{tab:speed_exp} 
% \vspace{-7mm}
% \begin{center}
% \scalebox{0.8}{\begin{tabular}{c|c|c}
% \Xhline{1pt}
% {\bf SC3D} \cite{2019sc3d} & {\bf P2B } \cite{2020p2b} & {\bf PTTR (Ours)} \\ 
% \hline\hline
% 66.3ms  & 23.6ms & \textbf{19.9ms} \\
% \Xhline{1pt}
% \end {tabular}}
% \vspace{-5mm}
% \end{center}
% \end {table}

% \smallskip
% \noindent\textbf{Inference Time.} Speed is a key factor in object tracking tasks. Hence, we test the model inference time on the KITTI test dataset with a Tesla V100 GPU. As reported in Table~\ref{tab:speed_exp}, under the same configurations, PTTR achieves the shortest average runtime of 19.9ms. 

\subsection{Analysis Experiments} \label{sec:ablations}

To evaluate the effectiveness of the components proposed in PTTR and PTTR++, we conduct extensive ablation studies on the KITTI \cite{kitti} dataset and report the experimental results.

\subsubsection{Ablation Studies on PTTR}

\begin{table}
\setlength{\tabcolsep}{4.75pt}
\caption {\textbf{Performance comparison on different point sampling methods.} D-FPS refers to distance-farthest point sampling and F-FPS denotes feature-farthest point sampling. RAS refers to our proposed Relation-Aware Sampling.} \label{tab:sample_experiment}
% Random random, D-FPS, F-FPS and our proposed Relation Sampling (RA)}. The performance is evaluated on the KITTI \cite{kitti} Tracking dataset.
\vspace{-4mm}
\begin{center}
\scalebox{0.8}{
\begin{tabular}{l|c|c|c|c|c}
\toprule[1.2pt]
{ Method} & { Car } & { Pedestrian} & { Van} & { Cyclist} & { Average} \\ 
\midrule
Random \cite{2020p2b} & \fontsize{8}{8}\selectfont 62.4 / 74.0 & \fontsize{8}{8}\selectfont 36.6 / 59.9 & \fontsize{8}{8}\selectfont 50.4 / 58.3 & \fontsize{8}{8}\selectfont 62.2 / 83.9 & \fontsize{8}{8}\selectfont 52.9 / 69.0 \\
D-FPS \cite{qi2017pointnet++}& \fontsize{8}{8}\selectfont 61.3 / 73.0 & \fontsize{8}{8}\selectfont 42.5 / 68.8 & \fontsize{8}{8}\selectfont 41.8 / 47.2 & \fontsize{8}{8}\selectfont 59.8 / 78.5 & \fontsize{8}{8}\selectfont 51.4 / 66.9 \\
F-FPS \cite{20203dssd} & \fontsize{8}{8}\selectfont 59.3 / 72.5 & \fontsize{8}{8}\selectfont 41.9 / 68.6 & \fontsize{8}{8}\selectfont 52.1 / 60.5 & \fontsize{8}{8}\selectfont 63.8 / 83.8 & \fontsize{8}{8}\selectfont 54.3 / 71.4 \\
\midrule
 RAS (Ours) & \fontsize{8.5}{8.5}\selectfont {65.2 / 77.4} & \fontsize{8.5}{8.5}\selectfont {50.9 / 81.6} & \fontsize{8.5}{8.5}\selectfont {52.5 / 61.8} & \fontsize{8.5}{8.5}\selectfont {65.1 / 90.5} & \fontsize{8.5}{8.5}\selectfont {58.4 / 77.8}\\
\bottomrule[1.2pt]
\end {tabular}
}
\end{center}
% \vspace{-2mm}
\end {table}

% \smallskip
\noindent\textbf{Point Sampling Methods.} 
We compare our proposed Relation-Aware Sampling (RAS) method with existing sampling approaches, including random sampling \cite{2020p2b}, distance-farthest point sampling (D-FPS) \cite{qi2017pointnet++} and feature-farthest point sampling (F-FPS) \cite{20203dssd}. 
As shown in Table~\ref{tab:sample_experiment}, RAS yields the best performance with a clear margin. 
% As we can observe, 
By utilizing RAS, our method achieves an increase of 5.5/8.8 in success/precision as compared to the random sampling baseline.
Small objects usually consist of fewer points, and hence are more sensitive to the point sparsity challenge. 
% It is worth noting that 
For the pedestrian class, which is the class of the smallest object size,
RAS significantly boosts the results from 36.6/59.9 to 50.9/81.6.

\begin {table}[t]
\setlength{\tabcolsep}{6pt}
\begin{center}
\caption{\textbf{Ablation studies on model components of PTTR.} For experiments that disable the Point Relation Transformer (PRT), we replace PRT with cosine similarity for feature correlation extraction as in existing methods \cite{2020p2b, 2019sc3d, zhou2021structure}. We also compare the performance w/ or w/o the Prediction Refinement Module (PRM).}
\vspace{-4mm}
\scalebox{0.8}{
\begin{tabular}{c|c|c|c|c|c|c}
% \Xhline{1pt}
\toprule[1.2pt]
{ PRT} & { PRM} & { Car } & { Pedestrian} & { Van} & { Cyclist}  & { Average}  \\ 
\midrule
 & & 40.2 / 52.0 & 23.0 / 41.6 & 25.9 / 34.7 & 30.0 / 57.8 & 29.8 / 46.5 \\ 
 \checkmark & & 62.9 / 74.3 & 49.1 / 77.7 & 50.7 / 58.7 & 64.1 / 90.0 & 56.7 / 75.2  \\ 
  & \checkmark  & 60.6 / 73.1 & 39.2 / 66.9 & 43.5 / 48.9 & 58.7 / 87.2 & 50.5 / 69.0\\ 
  \checkmark & \checkmark  & 65.2 / 77.4 & 50.9 / 81.6 & 52.5 / 61.8 & 65.1 / 90.5 & 58.4 / 77.8 \\ 
% \Xhline{1pt}
\bottomrule[1.2pt]
\end {tabular}
}
\label{tab:ablation_components} 
\end{center}
% \vspace{-4mm}
\end {table}

\begin{figure}[t]
    % \vspace{+3mm}
    \centering
    \includegraphics[width=1.0\linewidth]{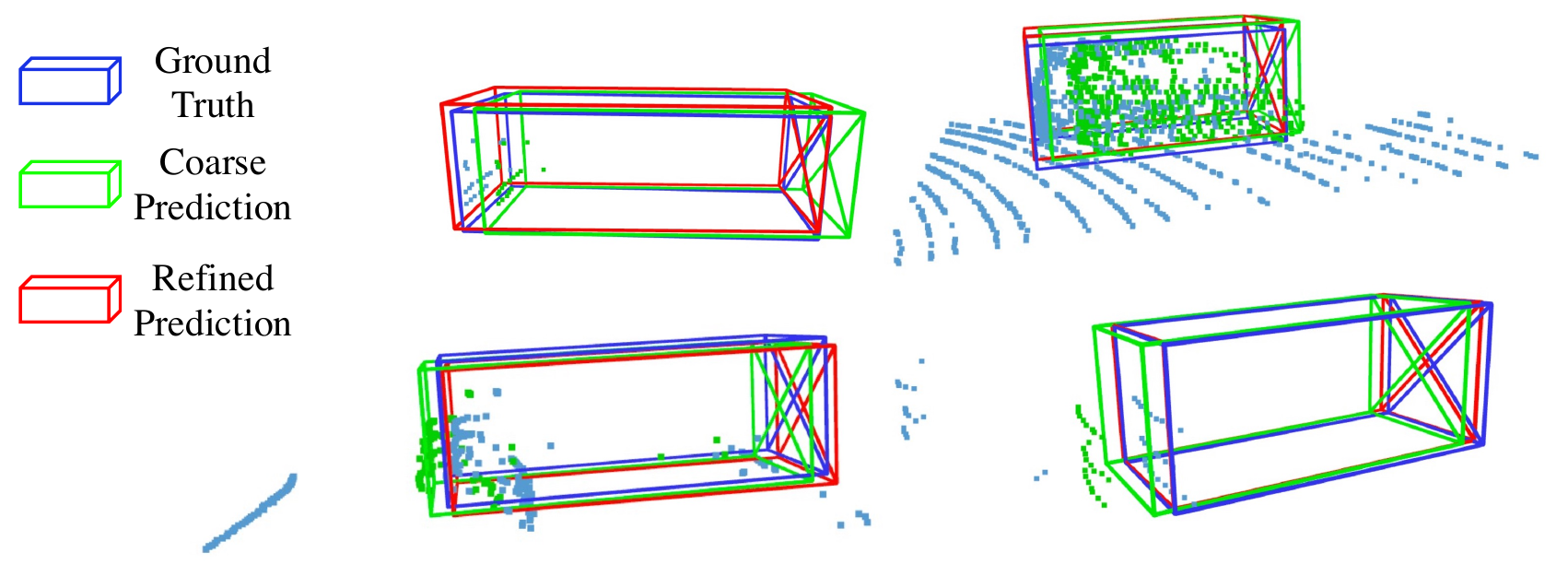}
    \vspace{-5mm}
    \caption{\textbf{Visualization of prediction refinement.}  
    We present 4 different objects to show that the proposed Prediction Refinement Module further corrects the coarse predictions through local feature pooling.}
    \label{fig:second_stage}
    % \vspace{-3mm}
\end{figure}

\smallskip
\noindent\textbf{Model Components.} 
We conduct experiments to investigate the effectiveness of the proposed Point Relation Transformer (PRT) and Prediction Refinement Module (PRM). 
For the ablation studies on PRT, we replace PRT with cosine similarity for feature correlation computation as in existing methods \cite{2020p2b, zhou2021structure, 2019sc3d}. 
We evaluate with the coarse prediction in experiments w/o PRM. 
As shown in Table~\ref{tab:ablation_components}, when both PRT and PRM are disabled, the performance degrades sharply from 58.4 to 29.8 in Success. 
Both PRT and PRM improve the model performance significantly compared to the base case. 
Note that replacing our PRT with cosine similarity will decrease the performance by 8.8 on average in terms of Success, which shows the strong feature matching capability of the PRT module. 
In addition, the two components are also complementary to each other that the best results are obtained when both are enabled. We also visualize the effect of the PRM in Fig.~\ref{fig:second_stage}. It can be observed that the coarse predictions are further corrected in the refinement stage, especially when point sparsity or large movements are present.

% We highlight that even without prediction refinement, our model still achieves the best average performance (56.7/75.2).
%  as compared to the best existing method (44.0/61.3).

\begin {table}[t]
 \setlength{\tabcolsep}{4.5pt}
\begin{center}
\caption{\textbf{Ablation studies on Relation Attention.} We investigate the effectiveness of the two major modifications in our proposed Relation Attention module. \textit{Offset} refers to offset attention and \textit{Norm} refers to feature normalization.}
\vspace{-2mm}
\scalebox{0.8}{
\begin{tabular}{c|c|c|c|c|c|c}
% \Xhline{1pt}
\toprule[1.2pt]
{ Offset} & { Norm} & { Car } & { Pedestrian} & { Van} & { Cyclist} & { Average}  \\ 
\midrule
 & & 55.4 / 68.0 & 36.6 / 65.1 & 34.6 / 38.6 & 55.9 / 78.8 & 45.6 / 62.6 \\ 
 \checkmark & & 56.6 / 69.1 & 40.3 / 67.3 & 48.3 / 59.6 & 63.7 / 90.3 & 52.2 / 71.6 \\ 
  & \checkmark  & 63.7 / 75.3 & 47.1 / 73.5 & 53.0 / 60.4 & 64.1 / 89.5 & 57.0 / 74.7 \\ 
  \checkmark & \checkmark  & 65.2 / 77.4 & 50.9 / 81.6 & 52.5 / 61.8 & 65.1 / 90.5 & 58.4 / 77.8 \\ 
\bottomrule[1.2pt]
\end {tabular}
}
\label{tab:attention_components} 
\end{center}
% \vspace{-5mm}
\end {table}

\smallskip
\noindent\textbf{Components of Relation Attention.} The main differences between our proposed Relation Attention and regular transformer attention are the L2-normalization applied on query and key features and the offset attention.
Ablation studies on each component are reported in Table~\ref{tab:attention_components}. 
Both two operations
improve the model performance, especially the L2-normalization.
It reveals that the cosine distance facilitates 
point cloud feature matching.

\subsubsection{Ablation Studies on PTTR++} \label{sec:ablation_pttr++}

\begin{figure*}[t]
    \centering
    \includegraphics[width=1.0\linewidth]{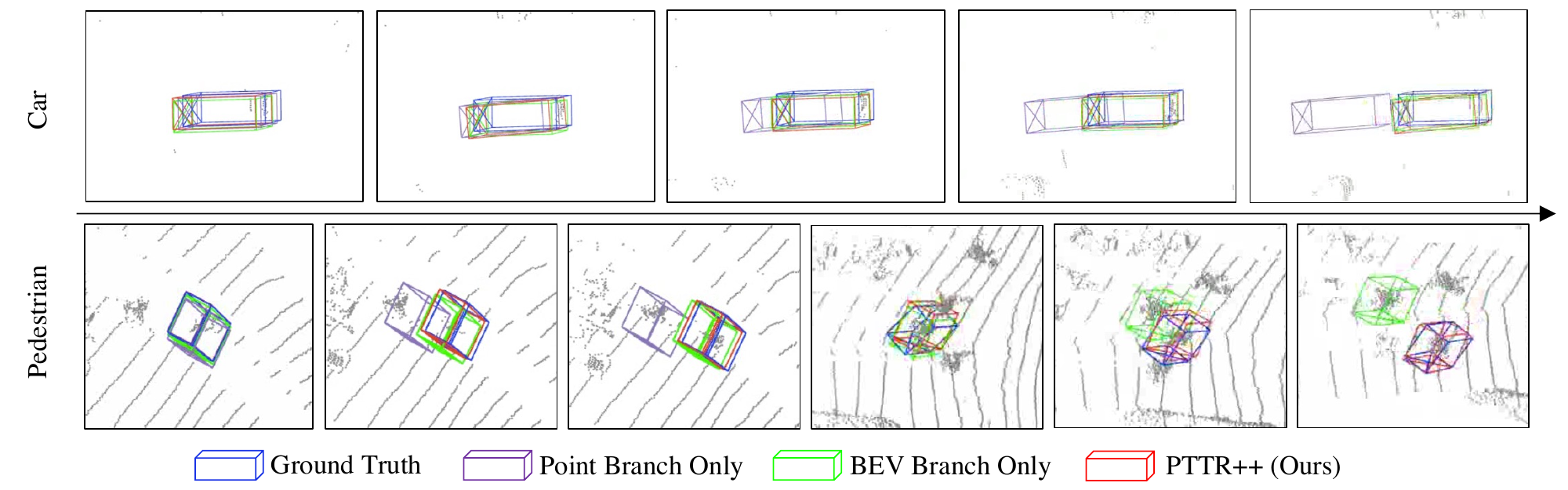}
    % \vspace{-12mm}
    \caption{\textbf{Visualization of tracking results from individual branches}. Our proposed PTTR++ generates more robust tracking predictions as compared to the point branch and the BEV branch by exploiting the complementary information of both representations. }
    \label{fig:visual_comparision}
    % \vspace{-2mm}
\end{figure*}

\noindent\textbf{Effectiveness of Point-BEV Fusion.}
To study the effectiveness of our proposed Point-BEV fusion, we conduct experiments by removing the fusion process and generating predictions independently on each branch. As shown in Table~\ref{tab:ablation_branch}, by utilizing the information from both branches, our proposed method outperforms the performance of each branch by a large margin, which validates the complementary effective of the point-wise and BEV representations and the effectiveness of our proposed fusion method. Moreover, it can be observed that the point branch and the BEV branch have large performance gaps on some of the classes (\eg, the success/precision of the BEV branch is 4.8/1.8 lower than the point branch on the pedestrian class), which implies that the two views have different advantages. We also qualitatively compare PTTR++ with the individual branches by visualizing their tracking results. As shown in Fig. \ref{fig:visual_comparision}, PTTR++ generates more robust and accurate tracking predictions by making use of the complementary information from the point-wise and BEV representations.

\begin{table}[t]
\setlength{\tabcolsep}{20pt}
\small
\caption {\textbf{Ablation studies on the effectiveness of Point-BEV fusion}.} \label{tab:ablation_branch} 
\vspace{-2mm}
\begin{center}\setlength{\tabcolsep}{4pt}{
\scalebox{0.75}{
\begin{tabular}{cc|c|c|c|c|c}
\toprule[1.2pt]
% {\textbf Point View} & {\textbf BEV} & {\textbf Car } & {\textbf Pedestrian} & {\textbf Van} & {\textbf Cyclist} & \textbf Average\\ 
Point &  BEV &  \multirow{2}{*}{Car} & \multirow{2}{*}{Pedestrian} &  \multirow{2}{*}{Van} & \multirow{2}{*}{Cyclist} &  \multirow{2}{*}{Average} \\
Branch & Branch &&&&& \\
\midrule
\checkmark &  &  62.9 / 74.3 & 49.1 / 77.7 & 50.7 / 58.7 & 64.1 / 90.0 & 56.7 / 75.2 \\
& \checkmark  &  68.5 / 80.7 &  44.3 / 75.9 &  51.3 / 59.4 &  70.1 / {93.4} &  58.6 / 77.4 \\ 
\checkmark & \checkmark  & {73.4} / {84.5} &  {55.2} / {84.7} &  {55.1} / {62.2} & {71.6} / 92.8 &  {63.8} / {81.0} \\
\bottomrule[1.2pt]
\end {tabular}}
}
% \vspace{-3mm}
\end{center}
\end {table}

\smallskip
\noindent\textbf{Point-BEV Fusion Methods.}
We experiment on different Point-BEV fusion methods including addition, global attention (\ie, attention with global average pooling), and point-wise attention. Besides, we perform feature fusion on the point branch and the BEV branch, respectively. As shown in Table~\ref{tab:ablation_fusion}, Point-BEV fusion via simple feature addition on either branch leads to significantly improved results as compared to the single-branch results reported in Table~\ref{tab:ablation_branch}, which demonstrates the complementary effect of the features from both views. By applying global attention during feature fusion, we observe slightly improved performance on the BEV branch and decreased performance on the point branch as compared to the feature addition fusion method. When point-wise attention is applied, we instead observe increased performance when fusion is performed on the point branch and lowered accuracy on the BEV branch. It shows that global attention is more suitable for BEV features, while point-wise attention is applicable to point features as different points might possess distinct feature responses and it is not advisable to apply unified channel-wise weights to all points. Finally, we select BEV-to-Point mapping and point-wise attention as our default setting since it achieves the best overall accuracy.

\begin{table}[t]
\setlength{\tabcolsep}{10pt}
\small
\caption{\textbf{Ablation studies on backbone design choices}. Downsample ratio refers to the number of times the BEV features are downsampled before the matching process.} \label{tab:ablation_bevbackbone} 
\vspace{-4mm}
\begin{center}\setlength{\tabcolsep}{4pt}{
\scalebox{0.78}{
\begin{tabular}{c|c|c|c|c|c}
\toprule[1.2pt]
 Downsample &  \multirow{2}{*}{Car} & \multirow{2}{*}{Pedestrian} &  \multirow{2}{*}{Van} & \multirow{2}{*}{Cyclist} &  \multirow{2}{*}{Average} \\ 
Ratio & & & & &  \\
\midrule
%  & Early & 69.8 / 80.7 & 50.2 / 79.4 &  51.5 / 58.5 & \textbf{73.7} / 93.0 &  61.3 / 77.9 & \textbf{16.4 ms} \\
% Mid & Early & 70.5 / 81.3 & 53.1 / 81.9 & 53.7 / 61.1 & 73.1 / \textbf{93.3}  & 62.6 / 79.4 & 20.3 ms\\
1x & {73.4}  / {84.5}  & {55.2}  / {84.7}  &  {55.1}  / {62.2}  &  {71.6} / {92.8} &  {63.8}  / {81.0}  \\
2x & 70.1 / 82.0 & 52.7 / 82.8 &  53.5 / 60.9 & 65.5 / 91.1 & 60.5 / 79.2 \\
4x & 67.3 / 79.1 & 49.9 / 77.1 & 50.4 / 56.3 & 69.4 / 92.6 & 59.3 / 76.3 \\
\bottomrule[1.2pt]
\end {tabular}}
}
% \vspace{-3mm}
\end{center}
\end {table}

\begin{table*}[t]
\setlength{\tabcolsep}{10pt}
\small
\caption{\textbf{Ablation studies on Point-BEV fusion methods}. } \label{tab:ablation_fusion}
% * Combined Attention refers to the use of the better performing point-wise attention for the point branch and global attention for the BEV branch.
\vspace{-4mm}
\begin{center}\setlength{\tabcolsep}{4pt}{
\scalebox{1.0}{
\begin{tabular}{c|c|c|c|c|c|c|c}
\toprule[1.2pt]
\multirow{2}{*}{Fusion Method} & \multicolumn{2}{c|}{Fusion Branch} & \multirow{2}{*}{Car} & \multirow{2}{*}{Pedestrian} & \multirow{2}{*}{Van} & \multirow{2}{*}{Cyclist} & \multirow{2}{*}{Average} \\ \cline{2-3}
& Point & BEV & & & & & \\ 
\midrule
\multirow{2}{*}{Addition} & \checkmark &  & \;73.0 / {84.5}\;  &	\;53.7 / 82.2\; &	\;54.8 / 61.9\; &	\;70.5 / 92.9\;	& \;63.0 / 80.4\;  \\ 
 & & \checkmark & 70.8 / 82.5  &	{56.7} / 84.8 &	51.5 / 58.6 & 71.6 / 93.6 &	62.7 / 79.9 \\ 
\midrule
% & \checkmark & \checkmark & 72.8 / 84.0 &	53.2 / 80.8 &	54.0 / 61.6 &	70.8 / 93.1 &	62.7 / 79.9 \\ \hline
\multirow{2}{*}{Global Attention} & \checkmark & &  71.5 / 83.3 &    52.0 / 81.1 &   54.3 / 60.5  &  69.8 / 92.7  & 61.9 / 79.4 \\ 
 & & \checkmark & 69.9 / 82.3 &		{56.7} / {85.4} &	51.6 / 58.1 & {74.1} / {94.3} &	63.1 / 80.0 \\ 
 \midrule
\multirow{2}{*}{Point-wise Attention} & \checkmark & &  {73.4} / {84.5} & 55.2 / 84.7 &  {55.1} / {62.2} &  71.6 / 92.8 &  {63.8} / {81.0}  \\ 
 & & \checkmark & 69.5 / 81.6 &	54.8 / 83.8 &	51.5 / 58.6 & 71.9 / 93.6 &	61.9 / 79.4 \\
%  Combined Attention\textsuperscript{*} & \checkmark & \checkmark & 71.8 / 84.3 &	53.0 / 81.3 &	55.3 / 62.7 & 70.3 / 92.8 &	62.6 / 80.3 \\
\bottomrule[1.2pt]
\end {tabular}}
}
% \vspace{-3mm}
\end{center}
\end {table*}

\smallskip
\noindent\textbf{BEV Branch Design Choices.} As introduced in Sec. \ref{method_matching}, the BEV branch differs from the point branch that we match template and search features before the BEV backbone network downsamples the feature maps. It allows BEV feature matching at a higher resolution for more accurate object localization. We compare the tracking accuracy when feature matching is performed under different resolutions. Apart from our default setting (no downsampling before feature matching), we apply layers in the backbone network to downsample the feature maps by 2x and 4x before the matching module and evaluate the performance. As shown in Table \ref{tab:ablation_bevbackbone}, matching the BEV features at higher resolution clearly leads to higher overall performance, which validates our proposed design choice.

\section{Limitation Discussion}
\begin{figure}[htb]
    % \vspace{+1mm}
    \centering
    \includegraphics[width=1.0\linewidth]{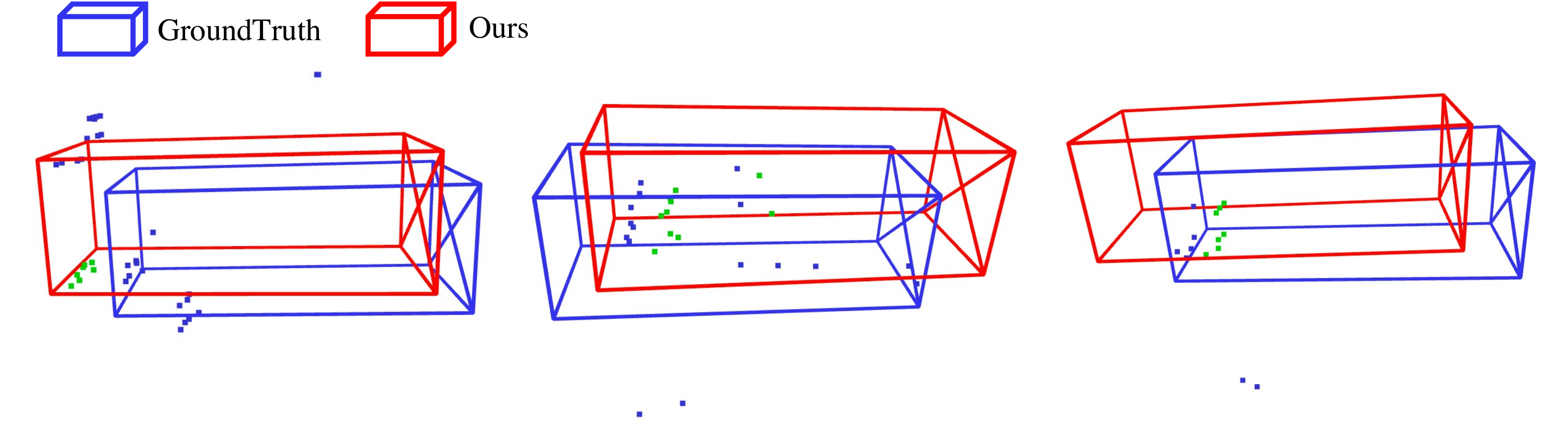}
    \vspace{-8mm}
    \caption{
    % \textbf{Failure cases of PTTR.} 
    \textbf{Examples of failure cases.} Our tracking failures mainly occur when the point clouds are too sparse.}
    \label{fig:failure}
    % \vspace{-3mm}
\end{figure}

We show in Fig.~\ref{fig:failure} the failure cases encountered by our model, which mainly occur when the point clouds are too sparse that the model can hardly capture enough patterns to effectively match template and search point clouds. One possible way to further mitigate this issue could be utilizing complementary multi-frame information for object tracking, which can be explored in future researches.

\section{Conclusion}   \label{sec:conclusion}

In this paper, we propose PTTR, a novel framework for 3D point cloud single object tracking, which contains a designed Relation-Aware Sampling strategy to tackle point sparsity, a novel Point Relation Transformer for feature matching, and a lightweight Prediction Refinement Module. Moreover, motivated by the advantages of the bird's-eye view of point cloud in capturing object motion, we design a more advanced framework named PTTR++ by exploiting the complementary information in point-wise and BEV representations. The proposed Point-BEV fusion strategy in PTTR++ substantially boosts the tracking performance and it can also be easily integrated with other tracking approaches as a generic method. 

\section*{Acknowledgement}
This study is supported under the RIE2020 Industry Alignment Fund – Industry Collaboration Projects (IAF-ICP) Funding Initiative, as well as cash and in-kind contribution from the industry partner(s).

% \ifCLASSOPTIONcompsoc
%   % The Computer Society usually uses the plural form
%   \section*{Acknowledgments}
% \else
%   % regular IEEE prefers the singular form
% \fi


\ifCLASSOPTIONcaptionsoff
  \newpage
\fi



\bibliographystyle{IEEEtran}
\bibliography{bib.bib}


\begin{IEEEbiography}[{\includegraphics[width=1in,height=1.25in,clip,keepaspectratio]{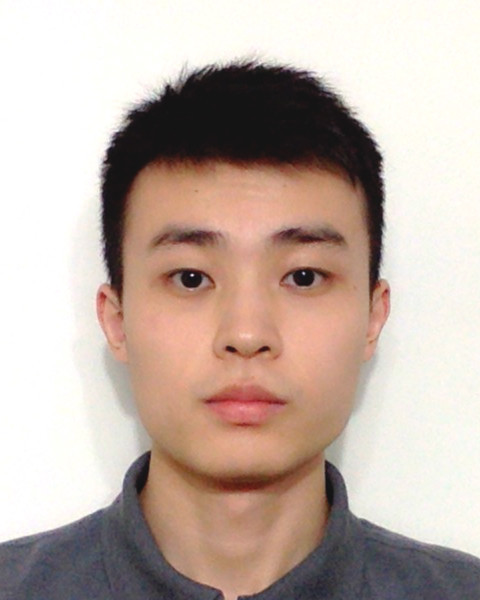}}]{Zhipeng Luo}
is currently a Ph.D. candidate with the School of Computer Science and Engineering, Nanyang Technological University, Singapore, under the supervision of Dr. Shijian Lu. He received his B.Eng. degree in mechanical engineering in 2015 and M.Sc. degree in computing in 2018 from National University of Singapore. He has  published multiple top conference papers in the field of computer vision. His research interests include computer vision, object detection, and object tracking.
\end{IEEEbiography}
\vspace{-10mm}

\begin{IEEEbiography}[{\includegraphics[width=1in,height=1.25in,clip,keepaspectratio]{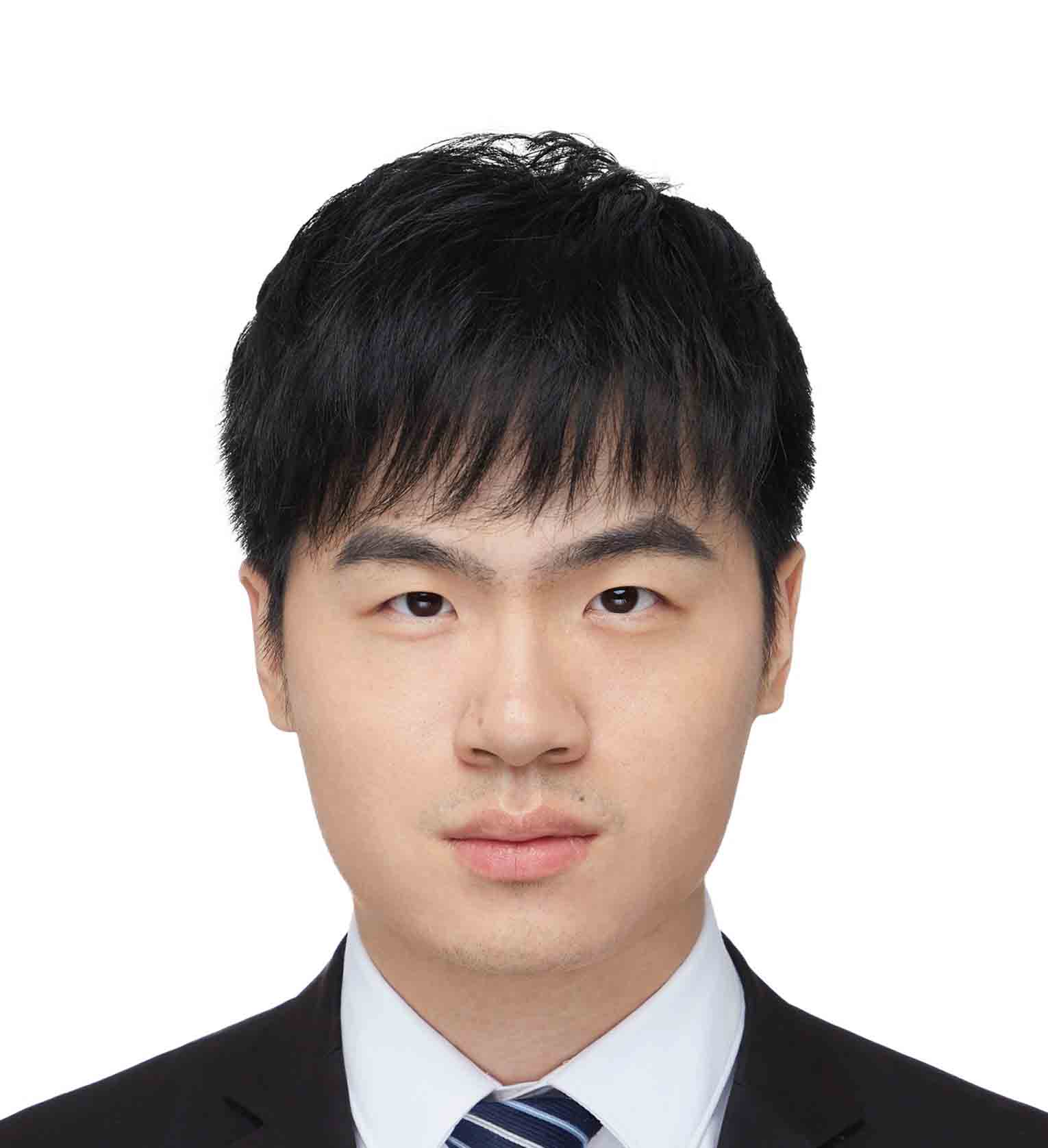}}]{Changqing Zhou}
is currently a research scientist at SenseTime. He received his B.Eng. degree  in Shanghai JiaoTong University in 2018 and M.Sc. degree in Nanyang Technological University in 2021. He has published multiple top conference papers in the field of computer vision and his research interests include computer vision and image processing.
\end{IEEEbiography}

\vspace{-10mm}

\begin{IEEEbiography}[{\includegraphics[width=1in,height=1.25in,clip,keepaspectratio]{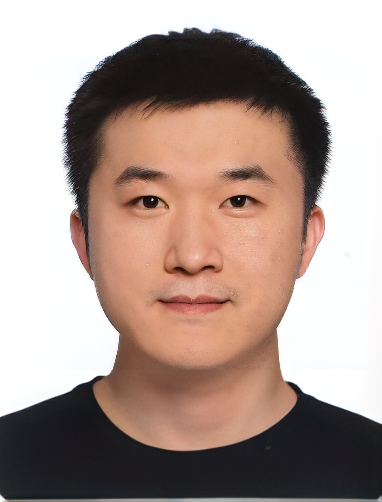}}]{Liang Pan}
		received the PhD degree in Mechanical Engineering from National University of Singapore (NUS) in 2019.
		He is currently a Research Fellow at S-Lab, Nanyang Technological University (NTU).  
		Previously, He is a Research Fellow at the Advanced Robotics Centre from National University of Singapore.
		His research interests include computer vision and
		3D point cloud, with focus on shape analysis, deep learning, and 3D human. 
		He also serves as a reviewer for top computer vision and robotics conferences, such as CVPR, ICCV, IROS and ICRA.
	\end{IEEEbiography}

\vspace{-6mm}

\begin{IEEEbiography}[{\includegraphics[width=1in,height=1.25in,clip,keepaspectratio]{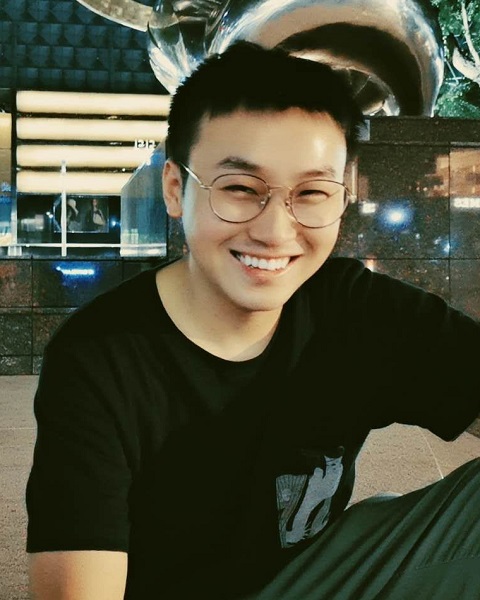}}]{Gongjie Zhang}
is currently working toward the Ph.D. degree in the School of Computer Science and Engineering, Nanyang Technological University, Singapore, under the supervision of Dr. Shijian Lu. He received his B.Eng. degree in electronic and information engineering in 2018 from Northeastern University, Shenyang, China. He has  published multiple journal and conference papers in the field of computer vision. He has also served as reviewer for several top journals and conferences such as T-PAMI, T-IP, CVPR, ICCV, and ACM\,MM. His research interests mainly include computer vision, object detection, few-shot learning, and meta-learning.
\end{IEEEbiography}

\vspace{-6mm}

\begin{IEEEbiography}[{\includegraphics[width=1in,height=1.25in,clip,keepaspectratio]{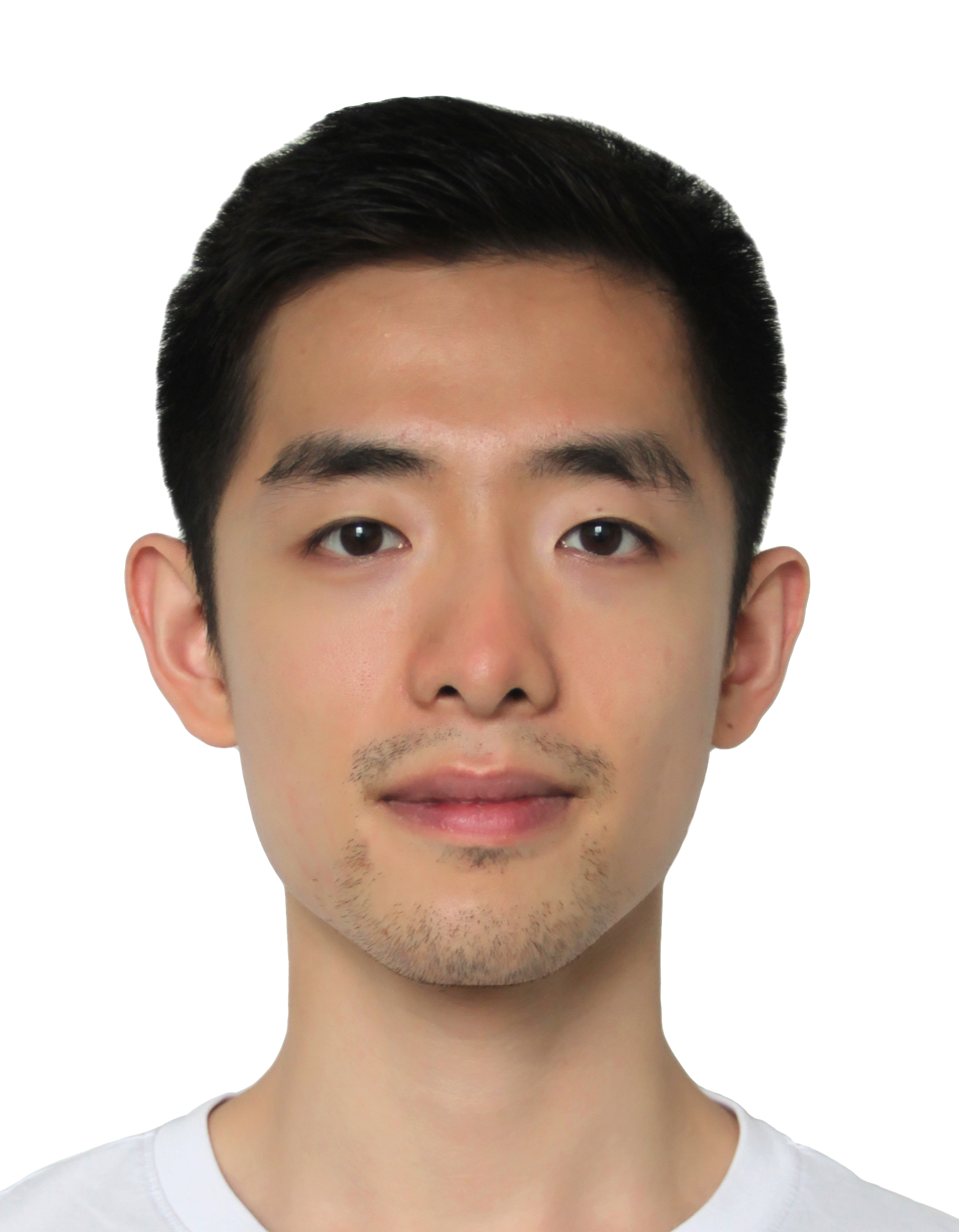}}]{Tianrui Liu}
is currently pursuing his Ph.D. degree at Nanyang Technological University. He received the B.Eng degree in Aerospace Engineering from Nanyang Technological University, Singapore, in 2018. His research interests include computer graphics and computer vision, especially in point cloud processing and augmented reality.
\end{IEEEbiography}

\vspace{-6mm}
\begin{IEEEbiography}[{\includegraphics[width=1in,height=1.25in,clip,keepaspectratio]{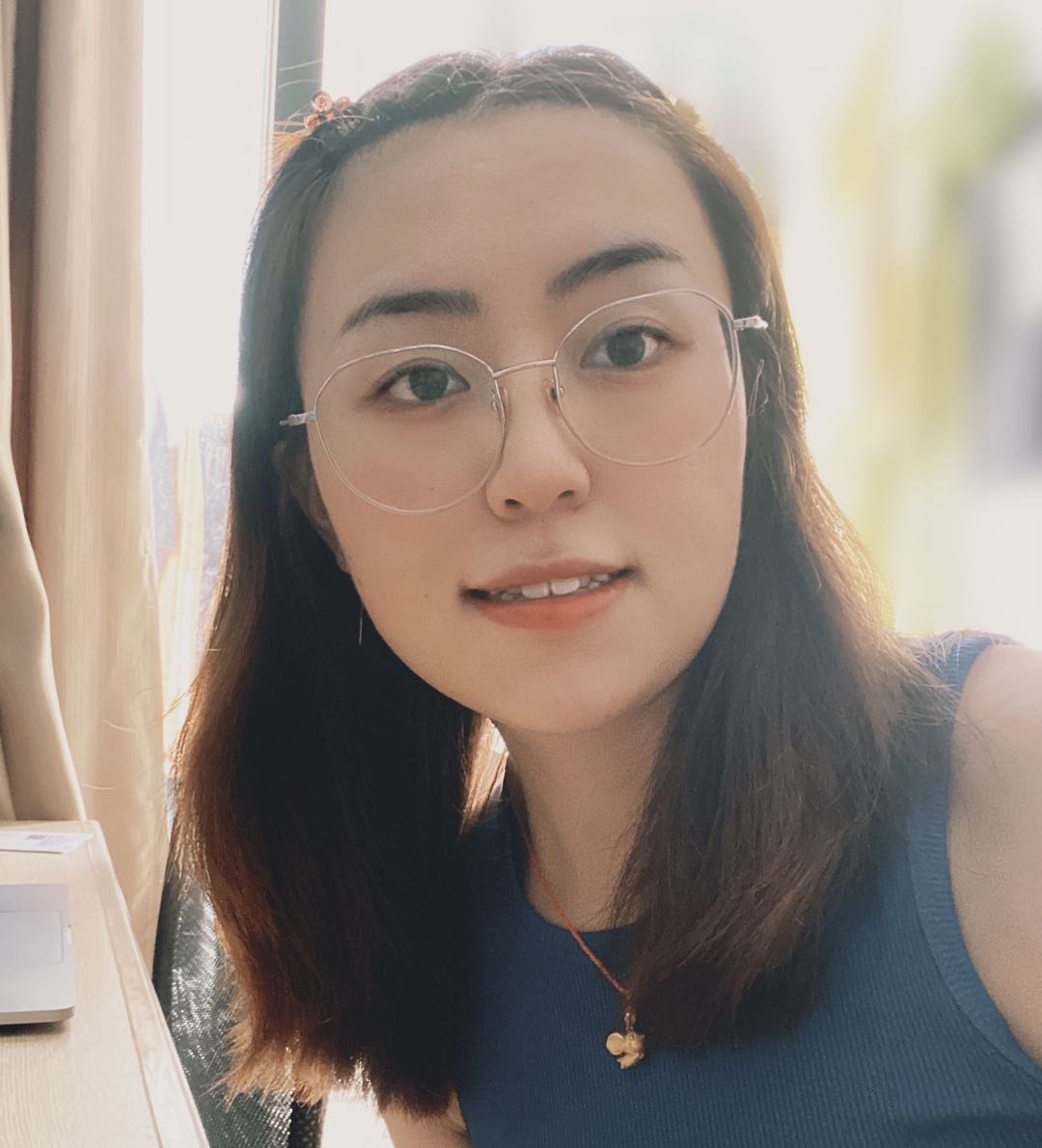}}]{Yueru Luo}
is currently a research assistant at the Chinese University of Hong Kong, Shenzhen. She received her M.Sc. degree from Nanyang Technological University in 2021. She is going to pursue a Ph.D. at CUHK, SZ and her research interests include computer vision, segmentation, and detection.
\end{IEEEbiography}

\vspace{-6mm}
\begin{IEEEbiography}[{\includegraphics[width=1in,height=1.25in,clip,keepaspectratio]{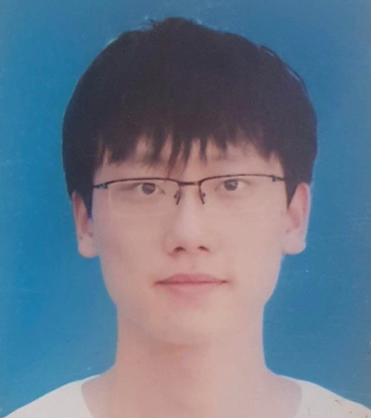}}]{Haiyu Zhao}
Haiyu Zhao is currently a senior researcher at SenseTime International Pte. Ltd. Previously, he was a research assistant at the Chinese University of Hong Kong Supervised by Prof. Xiaogang Wang. Before that, Haiyu received his bachelor degree at Beihang University. He has published over 15 papers (with more than 1,900 citations) on top-tier conferences in relevant fields, including CVPR, ICCV, ECCV, NeurIPS and ICLR.
\end{IEEEbiography}
\vspace{-6mm}

\begin{IEEEbiography}[{\includegraphics[width=1in,height=1.25in,clip,keepaspectratio]{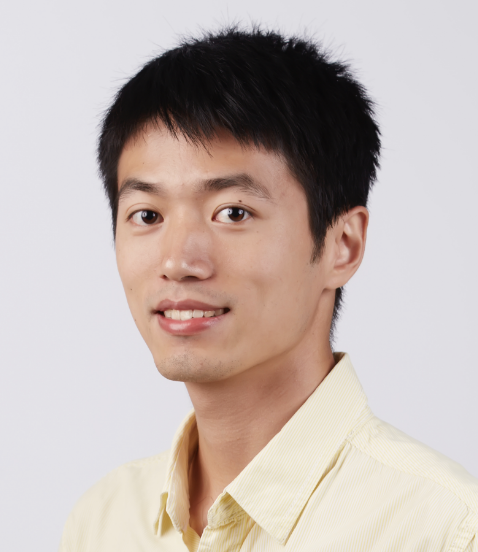}}]{Ziwei Liu}
is a Nanyang Assistant Professor at School of Computer Science and Engineering (SCSE) in Nanyang Technological University, with MMLab@NTU. Previously, he was a research fellow (2018-2020) in CUHK (with Prof. Dahua Lin) and a post-doc researcher (2017-2018) in UC Berkeley (with Prof. Stella X. Yu). His research interests include computer vision, machine learning and computer graphics.
Ziwei received his Ph.D. (2013-2017) from CUHK / Multimedia Lab, advised by Prof. Xiaoou Tang and Prof. Xiaogang Wang. He is fortunate to have internships at Microsoft Research and Google Research.
His works include Burst Denoising, CelebA, DeepFashion, Fashion Landmarks, DeepMRF, Voxel Flow, Long Tail, Compound Domain, and Wildlife Conservation.
\end{IEEEbiography}
\vspace{-6mm}

\begin{IEEEbiography}[{\includegraphics[width=1in,height=1.25in,clip,keepaspectratio]{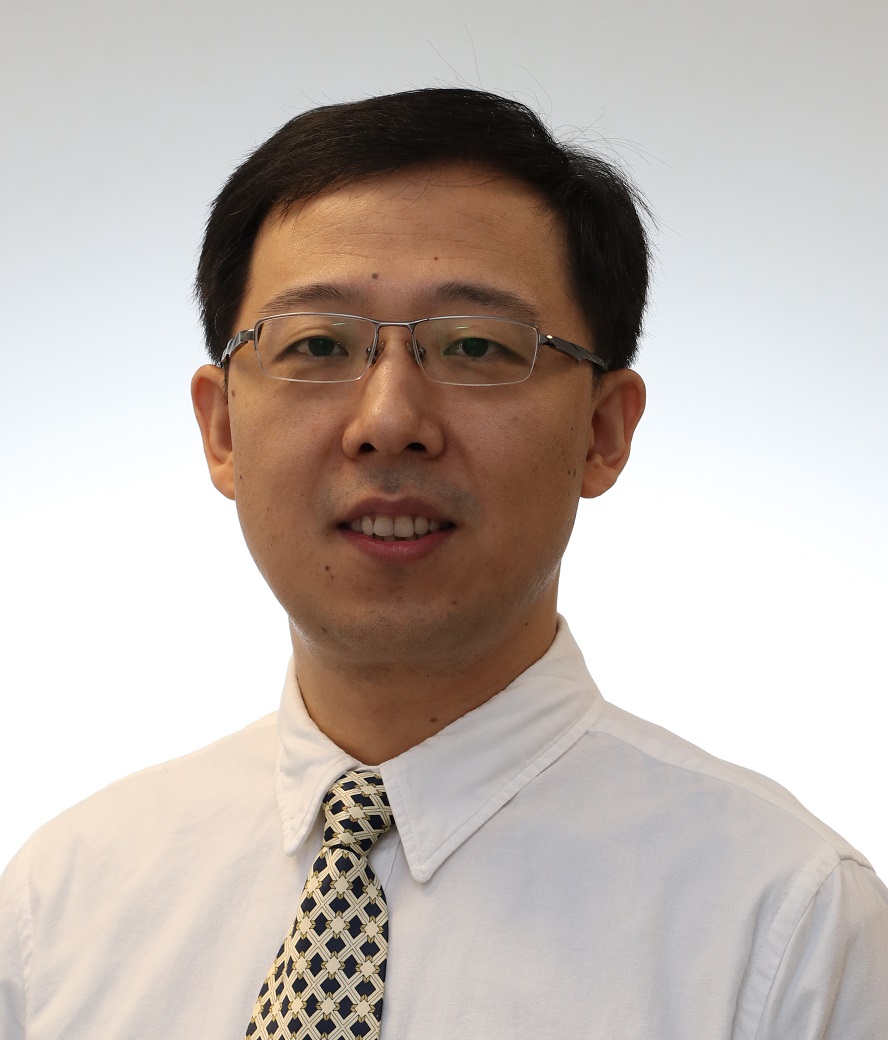}}]{Shijian Lu}
received his Ph.D. in electrical and computer engineering from National University of Singapore. He is currently an Associate Professor with the School of Computer Science and Engineering, Nanyang Technological University, Singapore. His major research interests include image and video analytics, visual intelligence, and machine learning. He has published more than 100 international refereed journal and conference papers and co-authored over 10 patents in these research areas. He is currently an Associate Editor for the journal Pattern Recognition (PR). He has also served in the program committee of a number of conferences, \eg, the Area Chair of the International Conference on Document Analysis and Recognition (ICDAR) 2017 and 2019, the Senior Program Committee of the International Joint Conferences on Artificial Intelligence (IJCAI) 2018 and 2019, \textit{etc}.
\end{IEEEbiography}


%\vfill


\end{document}